\documentclass{article}

\usepackage[dvipsnames]{xcolor}
\usepackage{microtype}
\usepackage{graphicx}
\usepackage{subfigure}
\usepackage{booktabs} 
\usepackage{xcolor}
\usepackage{float}
\usepackage{placeins}
\usepackage{afterpage}
\usepackage{tikz}
\usepackage{grffile}
\usepackage{subcaption}
\usetikzlibrary{shadows}

\usepackage{listings}
\usepackage{color}

\definecolor{codegreen}{rgb}{0,0.6,0}
\definecolor{codegray}{rgb}{0.5,0.5,0.5}
\definecolor{codepurple}{rgb}{0.58,0,0.82}
\definecolor{backcolour}{rgb}{0.95,0.95,0.92}

\lstdefinestyle{mystyle}{
    backgroundcolor=\color{white},   
    commentstyle=\color{codegreen},
    keywordstyle=\color{magenta},
    numberstyle=\tiny\color{codegray},
    stringstyle=\color{codepurple},
    basicstyle=\ttfamily\footnotesize,
    breakatwhitespace=false,         
    breaklines=true,                 
    captionpos=b,                    
    keepspaces=true,                 
    numbers=left,                    
    numbersep=5pt,                  
    showspaces=false,                
    showstringspaces=false,
    showtabs=false,                  
    tabsize=2
}

\usepackage{hyperref}

\usepackage[accepted]{icml2024}

\usepackage{amsmath}
\usepackage{amssymb}
\usepackage{mathtools}
\usepackage{amsthm}

\usepackage[capitalize,noabbrev]{cleveref}

\theoremstyle{plain}

\theoremstyle{definition}

\theoremstyle{remark}

\usepackage[textsize=tiny]{todonotes}

\icmltitlerunning{Just Cluster It}

\begin{document}

\twocolumn[
	\icmltitle{Just Cluster It: An Approach for Exploration in High-Dimensions\\ using Clustering and Pre-Trained Representations}



	\icmlsetsymbol{equal}{*}

	\begin{icmlauthorlist}
		\icmlauthor{Stefan Sylvius Wagner}{yyy}
		\icmlauthor{Stefan Harmeling}{xxx}
		
		
	\end{icmlauthorlist}

	\icmlaffiliation{yyy}{Department of Computer Science, Heinrich Heine University Düsseldorf, Germany}
	\icmlaffiliation{xxx}{Department of Computer Science, Technical University Dortmund, Germany}


	\icmlcorrespondingauthor{Stefan Wagner}{stefan.wagner@hhu.de}

	\icmlkeywords{Machine Learning, ICML}

	\vskip 0.3in
]



\printAffiliationsAndNotice{}  

\begin{abstract}
	In this paper we adopt a representation-centric perspective on exploration in reinforcement learning, viewing exploration fundamentally as a density estimation problem.
	We investigate the effectiveness of clustering representations for exploration in 3-D environments, 
	based on the observation that the importance of pixel changes between transitions is 
	less pronounced in 3-D environments compared to 2-D environments,
	where pixel changes between transitions are typically distinct and significant.
	We propose a method that performs episodic and global clustering on random representations and on pre-trained DINO representations to count states, i.e, estimate pseudo-counts. 
	Surprisingly, even random features can be clustered effectively to count states in 3-D environments, 
	however when these become visually more complex, pre-trained DINO representations are more effective thanks to the pre-trained inductive biases in the representations. 
	Overall, this presents a pathway for integrating pre-trained biases into exploration. We evaluate our approach on the VizDoom and Habitat environments, 
	demonstrating that our method surpasses other well-known exploration methods in these settings.
\end{abstract}

\section{Introduction}
\label{sec:introduction}

\begin{figure*}
    \centering
    \subfigure[]{

	\begin{tikzpicture}	

		\node[draw, thick, color=black, fill=white, rounded corners=0.0mm, inner sep=1mm, drop shadow] {

		\begin{tikzpicture}[scale=0.4]
				\node[inner sep=0pt] (img1) at (0,0) {\includegraphics[width=1.5cm]{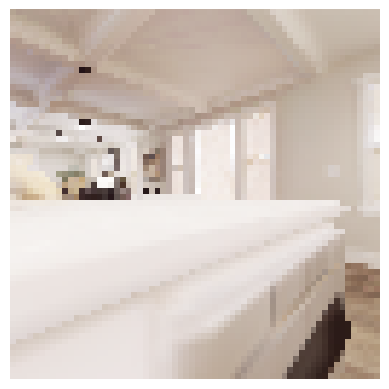}};
			
				\node at (2.5,0) {$\boldsymbol{-}$};
			
				\node[inner sep=0pt] (img2) at (5,0) {\includegraphics[width=1.5cm]{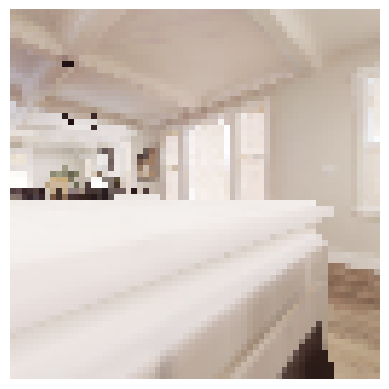}};
			
				\node at (7.5,0) {$\boldsymbol{=}$};
			
				\node[inner sep=0pt] (img3) at (10,0) {\includegraphics[width=1.5cm]{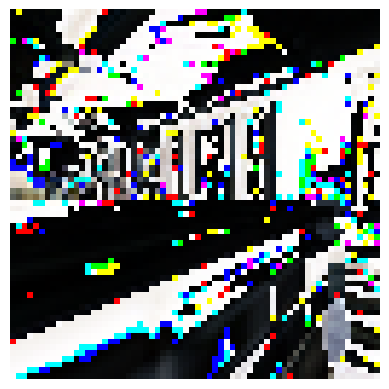}};
				\node[inner sep=0pt] (img4) at (0,-4.5) {\includegraphics[width=1.5cm]{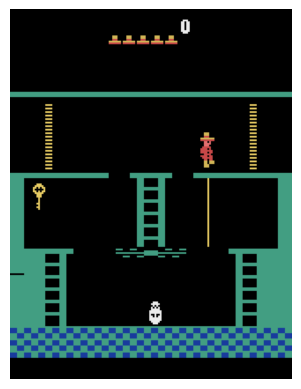}};
			
				\node at (2.5,-4.5) {$\boldsymbol{-}$};
			
				\node[inner sep=0pt] (img5) at (5,-4.5) {\includegraphics[width=1.5cm]{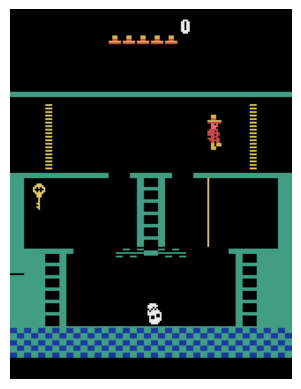}};
			
				\node at (7.5,-4.5) {$\boldsymbol{=}$};
			
				\node[inner sep=0pt] (img6) at (10,-4.5) {\includegraphics[width=1.5cm]{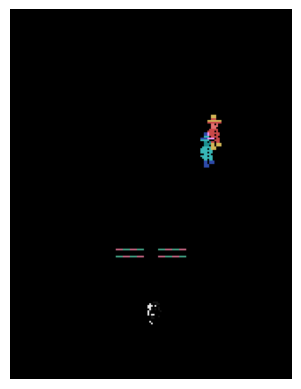}};
			\end{tikzpicture}
		};
		\node[text width=4.5cm, text centered, above of=img1] at (0, 2.7) {\fontfamily{pcr}\selectfont\footnotesize{\textcolor{blue}{In 3-D Exploration:} Pixel change magnitude high, but saliency low.}};	
		\node[text width=5.5cm, text centered, above of=img1] at (0, 1.8) {\fontfamily{pcr}\selectfont\footnotesize{\textcolor{red}{Problem:} Bad intrinsic reward}};	

		\node[text width=5cm, text centered, above of=img1] at (0, 1.3) {\fontfamily{pcr}\selectfont\footnotesize\textbf{Solution? \textcolor{green}{Just Cluster It!}}};	

	\end{tikzpicture}
	\label{fig:pixelchanges}
	} \hfill
    \subfigure[]{ 		
		\begin{tikzpicture}
			\node[draw, thick, color=black, fill=white, rounded corners=0.0mm, inner sep=1mm, drop shadow] {

				\begin{tikzpicture}[scale=0.4]

				\node[inner sep=0pt] at (0,0) {\includegraphics[width=.08\textwidth]{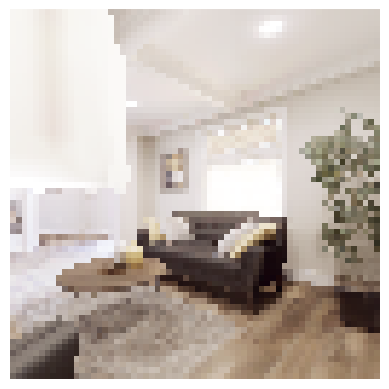}};
				\node[inner sep=0pt] at (3.2,0) {\includegraphics[width=.08\textwidth]{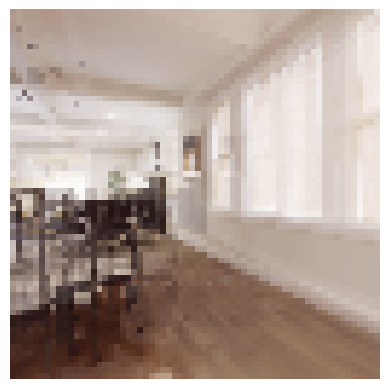}};
				\node[inner sep=0pt] at (6.4,0) {\includegraphics[width=.08\textwidth]{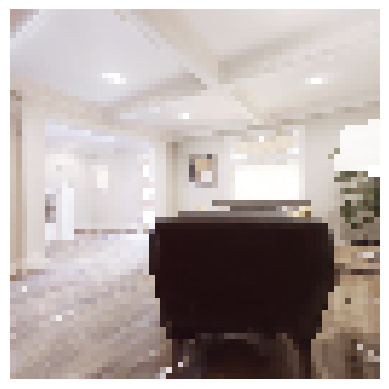}};
			
				\node[inner sep=0pt] at (0,-3.5) {\includegraphics[width=.08\textwidth]{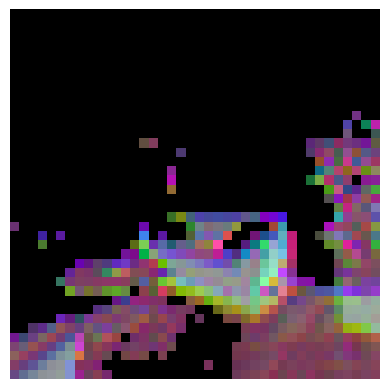}};
				\node[inner sep=0pt] at (3.2,-3.5) {\includegraphics[width=.08\textwidth]{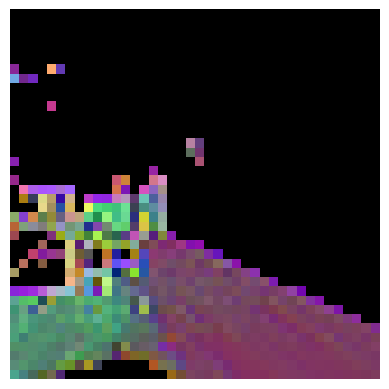}};
				\node[inner sep=0pt] at (6.4,-3.5) {\includegraphics[width=.08\textwidth]{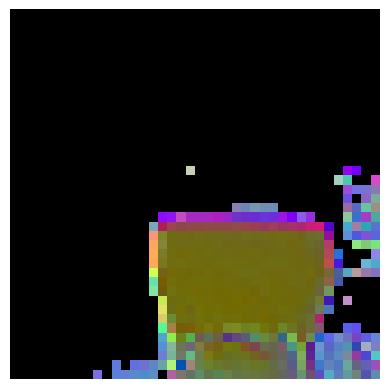}};

				\end{tikzpicture}
			};
			\node[text width=4cm, text centered, above of=img1] at (0, 1.3) {\fontfamily{pcr}\selectfont\footnotesize{Cluster pre-trained embeddings for state aggregation}};	

		\end{tikzpicture}
		\label{fig:dino_features}
	}\hfill
	\subfigure[]{
		\begin{tikzpicture}
			\node[draw, thick, color=black, fill=white, rounded corners=0.0mm, inner sep=1mm, drop shadow] {
				\includegraphics[width=0.35\textwidth]{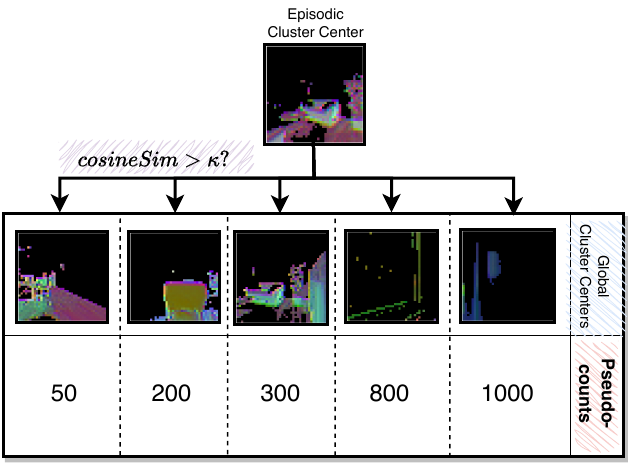}
			};
			\node[text width=5cm, text centered, above of=img1] at (0, 2.2) {\fontfamily{pcr}\selectfont\footnotesize{Combine episodic and global clustering to estimate pseudo-counts}};	
		\end{tikzpicture}
		\label{fig:cluster_table}
	} 
    \caption{\textbf{``Just Cluster It'' in a nutshell:} \textbf{(a)} Density estimation for exploration in 
	3-D environments is challenging, because the magnitude of pixel change is large, but the saliency of a single transition is low.
	This is in contrast to 2-D environments where every transition is distinct and therefore salient. Therefore, we propose clustering
	representations on an episodic and global level to estimate accurate pseudo-counts that reflect the state-space distribution.
	\textbf{(b)} We show pre-trained representations from a DINO model. The embeddings from the DINO model are able to
	extract relevant features from the observations, which is useful for clustering. The bottom images are thresholded embeddings.  \textbf{(c)} To estimate pseudo-counts, we store episodic cluster centers in a global cluster table 
	over time by matching episodic cluster centers with previously added cluster centers in the cluster table. A new cluster center is added to the table, if the cosine similarity to existing cluster centers is below a threshold $\kappa$.}
\end{figure*}
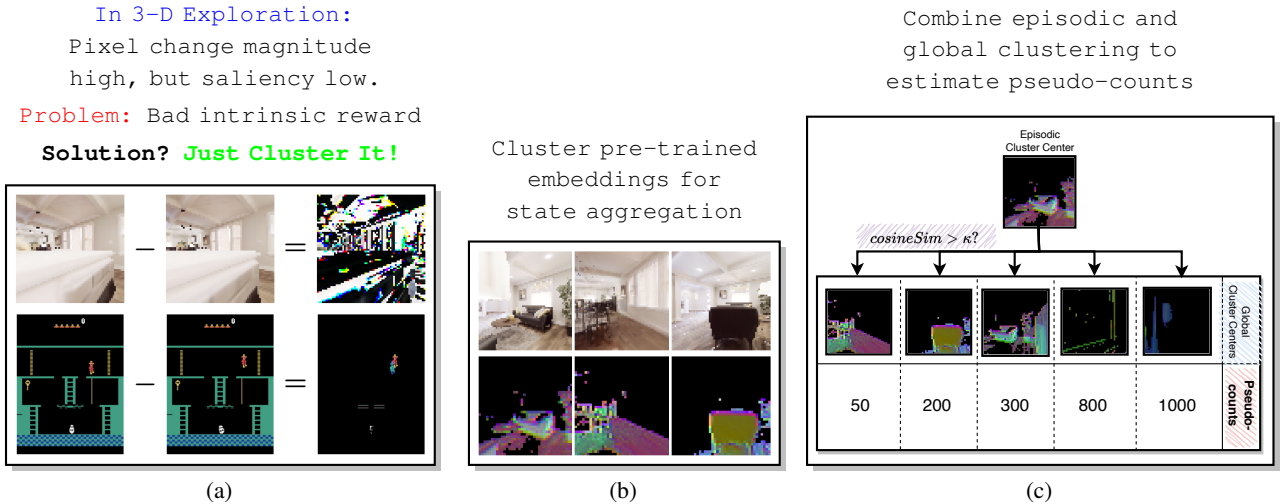

Exploration in reinforcement learning can inherently be seen as a density
estimation problem. Approximating the density of the state-space throughout
learning, allows the agent to determine which parts of the environment it has
visited. This paradigm is at the core of exploration via intrinsic motivation
where an intrinsic reward signal is created to quantify novelty. A simple way to
do this is by counting states \citep{2016TangHashtag}. However, simply counting
all states naively is not possible, since counts are very sparse in
high-dimensional or stochastic settings and the memory requirements to store
every distinct count for a given state are very high. There are several
approaches to estimating the state density: One way is to use some density
estimation model that given an observation assigns a pseudo-count \citep{2016BellemarePseudo}, where a pseudo-count 
is an estimated count for a state given a density model. Other methods
are based on using the prediction error, where the estimate for the state
density is based on whether a model can predict fixed target features provided
by a second model \citep{2017PathakCuriosity,2018BurdaRandom,2020RaileanuRIDE}. Established methods such as Random Network Distillation \citep{2018BurdaRandom} show
that novelty can be determined simply by calculating the prediction error of
random features. Different states will provide different feature vectors hence
the model learns an indirect density estimate of the state space.
Recent methods such as BYOL-Explore \citep{2022GuoBYOL} replace random features with trained features, with great
success.


\paragraph{The need for state aggregation in 3-D vs. 2-D.} 
Due to its inherent simplicity, counting states for exploration in non-tabular environments has been studied extensively \citep{2016BellemarePseudo, 2016TangHashtag,2017OstrovskiCountBased}. 
While counting states for exploration works well in tabular settings, it becomes intractable in high dimensional settings if  
every state is counted explicitly. \citet{2016BellemarePseudo} were among the first to propose the estimation of pseudo-counts, where 
the counts are modeled by some density model of the state-space. Essentially, this transforms the problem of exploration into a density estimation problem of the 
state-space. 
A central property for a density model in exploration 
is that the model should be able to distinguish different states, 
such that salient transitions can be identified in the intrinsic reward.
In the case of counting states, the states should be aggregated in a sensible way such that the counts are not too granular, but also not too coarse \citep{2017OstrovskiCountBased,2021EcoffetFirstReturn}.
We identify a particular phenomenon regarding the importance of pixel changes in 3-D environments that requires states to be aggregated in order to provide a salient intrinsic reward.
We provide a straightforward example in Figure \ref{fig:pixelchanges}, to demonstrate that state aggregation is necessary in 
3-D environments compared to 2-D enviroments and thus must be accounted for 
by a density model in 3-D environments. We characterize this via transition saliency of pixels, i.e., the significance of pixel changes between two observations.  
For instance, in 2-D environments pixel change between transitions is usually small, but the significance of the pixel change is large. 
As a consequence, almost every observation contains new information regarding the current state of an agent.
One example is Montezumas Revenge where individual groups of pixels change
substantially when an action is performed, since the playable character figure changes
its pixels when e.g., moving from left to right or when interacting
with an object.
In 3-D environments by contrast, although the magnitude of pixel change is large, these
pixel changes are usually not as signficant, since after a single transition
the relationship between the pixels in the observations remains largely the same.
Rather, a more effective reference point for estimating the state
density should be centered around salient features in the representations, such that
observations with similar features are considered similar. 
In conclusion, in 2-D enviroments the transition saliency of pixels is high, whereas in 3-D environments the transition saliency of pixels is low.
Thus, the underyling density model must account for this lack of saliency in 3-D environments to yield a meaningful intrinsic reward.
Similar thoughts have also been presented by \citet{2021EcoffetFirstReturn,2017OstrovskiCountBased}.
We argue that tackling this transition saliency problem may bridge the gap from count-based methods to prediction-error-based methods which excel at providing a salient intrinsic reward signal when
observations are discrete and pixel changes are meaningful.
However, they perform worse when
the observations are high-dimensional and individual transitions are not as
meaningful.  
Overall, if an appropriate density model can be found to estimate pseudo-counts in high dimensional observations, we can 
exploit the favorable theoretical guarantees of count-based methods \citep{2016BellemarePseudo} and avoid catastrophic forgetting in neural networks.
With this in mind, we deduce that a clustering-based approach, which is able to find the right level of state aggregation by finding 
shared features in the observations, may be much more effective in high-dimensional environments, 
provided the underlying representations are expressive enough. Ideally, the level of state aggregation should 
be determined from the underlying features contained in the observations.
\paragraph{Exploration through Episodic and Global Clustering of High-Dimensional Representations.} 
To this end, we propose a method for exploration that performs episodic and global clustering of (pre-trained) representations to
estimate pseudo-counts. More precisely, in each episode we fit a Gaussian
mixture model on random features or on features from a pre-trained
DINO \citep{2023OquabDINOv2,2021CaronDINO} model to exploit the invariances contained in the representations. The Gaussian mixture model determines
cluster means, i.e., centers for the
observations in the current episode, which can then be compared with 
cluster centers previously stored in memory (see Figure \ref{fig:cluster_table}).
Particularly when utilizing pre-trained representations, as demonstrated in Figure \ref{fig:dino_features}, 
we see that these features align well with objects in the scene. We consider these features as various modes in the observational distribution, 
which makes them well suited for a Gaussian mixture model.
The global clustering is then done in the cluster table, where
a cosine similarity threshold $\kappa$ is used to determine whether a new episodic cluster center should be added to memory or not.
This is akin to the stick breaking process in Dirichlet processes and thus allows us to control the granularity of the
inter-episodic clustering. We then calculate the intrinsic reward for each observation in the episode as the inverse square root of the pseudo-counts \citep{Strehl2008Counts}.
We conduct experiments on both the VizDoom \citep{Kempka2016ViZDoom} and Habitat \citep{habitat19iccv,szot2021habitat,puig2023habitat3} environments, demonstrating that 
even features chosen at random outperform other exploration methods when dealing 
with 3-D environments. 
\textbf{Contributions}:
\textbf{(i)} We introduce an exploration method that determines novelty by estimating pseudo-counts 
	of clustered pre-trained representations. We propose a method that involves both episodic and global clustering. 
	The episodic clustering aggregates similar representations by fitting a Gaussian mixture model to pre-trained DINO 
	representations or to random features for every episode. The global clustering is done over time by matching and 
	storing episodic cluster centers in a global cluster table that aggregates pseudo-counts.   
	Empirically, we demonstrate the effectiveness of this procedure in the chosen 3-D environments.

	\textbf{(ii)} We show that both random and pre-trained DINO features can be clustered effectively
	to aggregate states in 3-D environments and thus estimate pseudo-counts effectively. While random features work well on a simpler 3-D environment (VizDoom environment), 
	with real-world observations (Habitat environment),
	the priors contained in the pre-trained DINO representations improve performance substantially.

	\textbf{(iii)} By ablating two components of our method - the cosine similarity threshold $\kappa$ and the episodic clusters, 
	we show the effectiveness of clustering observations in high-dimensional 3-D environments.

\section{Background}
\label{sec:related_work}

\paragraph{Count based methods and estimating pseudo-counts.}
Pseudo-counts as defined by \citet{2016BellemarePseudo} stand at the center of generalizing counts
to high dimensional state spaces. They define state counts relative to density model that approximates the underlying
state-space distribution. In principle, this definition of
pseudo-counts is agnostic to the given state representation \citep{2017OstrovskiCountBased}. There are various
exploration methods that rely on state counting, either to detect
repeating observations or in order to normalize the intrinsic reward 
\citep{2020RaileanuRIDE,Zhang2021NovelD,2023WagnerCyclophobic,parisi2021interesting,2016TangHashtag}. 
These methods ultimately propose pseudo-counts when extending
to high-dimensional state-spaces. \citet{2016BellemarePseudo} also demonstrate that estimating
pseudo-counts using a density model provides a measure of either information gain, i.e.,
prediction gain. This shows that pseudo-counts are related to intrinsic
reward methods that utilize prediction error as an intrinsic reward signal. When calculating 
an intrinsic reward bonus based on pseudo-counts, a common choice to calculate the intrinsic bonus is
the inverse square root of the pseudo-counts as proposed by \citet{Strehl2008Counts}.

\paragraph{Novelty via prediction error.}
Most recent exploration methods fall under the category of exploration via
prediction error. The idea is that the prediction error of a model with
regard to some target features is a measure of novelty.
Various versions of these methods exist, like Random Network Distillation \citep{2018BurdaRandom} (RND),
in which the target features come from a neural network that is both fixed and initialized randomly.
\citet{2017PathakCuriosity}  learns inverse dynamics with a separate model that
learns to predict the correct actions between state transitions which makes the learned
representations invariant to uncontrollable parts of the enviroment. \citet{2020RaileanuRIDE} predict the target features of 
the next state given the current state instead of predicting the target features of the same state. 
Recent methods such as BYOL-Explore \citep{2022GuoBYOL} learn representations online with BYOL \citep{2020GrillBYOL}
leading to better target features.
Overall, these models estimate the density of
the state space through the target features they approximate.
Thus, the target features influence the state-space
density that is learned. 
An advantage of prediction-error-methods is that since they are parametrized by a neural network they can generate an intrinsic reward signal 
ad-hoc from raw observations. 

\paragraph{Density estimation via clustering representations.}
A basic method for estimating the density of the state-space involves the use of clustering.
\citet{2019MaClustered} cluster the observations directly by performing k-means on the observations.
Then they calculate the pseudo-counts on an episodic level
and do not store the counts throughout the entirety of learning.
``Never Give Up'' \citep[NGU, ][]{2020PuigdomenechBadiaNever} also make use of episodic clustering to estimate
episodic novelty. For each episode, NGU determines the sum of squared errors between the k-nearest neighbors of the current state and the current state itself.
To capture long-term dependencies a global intrinsic reward is calculated via random network distillation.
E3B \citep{2022HenaffE3B} performs episodic clustering by fitting a covariance matrix at each step and
computing eigenvalues of the updated covariance matrix. The scale of the
eigenvalues then determines the novelty.  Apart from episodic clustering
methods there are also methods that perform global clustering. Go-Explore \citep{2021EcoffetFirstReturn} maps
observations to cells, which can also be seen as a form of clustering. However, instead of performing explicit clustering, a heuristic is used to find 
suitable values for downscaling parameters such as width, height and depth. The idea is that similar observations are mapped to the same cell. 
RECODE \citep{2023SaadeUnlocking} attempts to estimate pseudo-counts by approximating clusters via Kernel Density Estimation (KDE). 
The current embedding is taken as central point and a density is fitted to find the nearest approximated cluster means stored in a table.
If a new embedding is significantly different from its neighbors in the table it replaces an old embedding. 
RECODE is closely related to our method, with its Dirichlet process inspired approach.
However, RECODE is also somewhat susceptible to the problem of transition saliency in 3-D environments, since it updates the cluster means at every step with only a single embedding as reference point.
We also simplify the stick breaking process for the cluster table, as we only have a single threshold $\kappa$ that determines whether a new cluster center should be added to the cluster table or not.


\section{Just Cluster It: Exploration via Episodic and Global Clustering}
In this section, we describe our exploration method ``Just Cluster It!'', which constists of a two-stage clustering:  first, we cluster after each episode, which we call \emph{episodic clustering}, and second, we collect the newly found clusters in a global table of cluster centers, which we call \emph{global clustering}.  The cluster centers of the global clustering maintain a list of pseudo-counts to approximate the state space distribution.
Therefore, it is important that the episodic cluster centers are representative of the observational distribution in the current episode, in order
to aggregate the pseudo-counts correctly. Over time, the cluster table will contain pseudo-counts for the whole state-space distribution.  Note that we call the counts \emph{pseudo}, since they are defined relative to a mixture model.

For each episode, we employ a two-step process:  first, we perform episodic clustering by fitting a Gaussian mixture model 
on the batch of representations of the current episode. This yields several episodic cluster centers. 
The second step updates the global cluster table that contains cluster centers from previous episodes and their corresponding counts.
For each episode, counts from episodic cluster centers are then either aggregated to existing cluster centers, or a new cluster center is added 
depending on a cosine similarity threshold.
This process of adding cluster centers continuously is similar to the stick breaking procedure in Dirichlet processes. 
However, instead of sampling weights from a beta distribution to determine the new cluster size, 
we simply add a new cluster center if the maximum cosine similarity is below a threshold $0\le\kappa\le 1$. In this sense, $\kappa$ controls the granularity of the clusters in the cluster table and is related to the concentration parameter 
for the Beta distribution in Dirichlet processes. 
The closer $\kappa$ is to one, the more cluster centers are added to the cluster table as shown in Figure \ref{fig:bandwidth}. 
The key insight resulting from our method is that (pre-trained) representations in high dimensions are expressive enough to be clustered effectively on an episodic level, while also contributing to the approximation of the global state space distribution.


\paragraph{Notation and Preliminaries.}
We define an MDP as a tuple
$(\mathcal{S}, \mathcal{A}, \mathcal{P}, \mathcal{R}, \gamma)$ with interactions at discrete time steps
$t=0,1,2,3, \ldots$. The agent receives a state from the environment
as a random variable $S_t \in \mathcal{S}$, where $S_t = s$ is some
representation of the state from the set of states $\mathcal{S}$ at timestep
$t$. The agent then selects an
action $A_t \in \mathcal{A}$ where $A_t = a$ is some action in the possible set
of actions $\mathcal{A}$ for the agent at timestep $t$. This action is selected
according to a policy ${\pi(a|s)}$ or $\pi(s)$ if the policy is
deterministic. One time step later after taking the action, the agent
receives a numerical reward which is a random variable $R_{t+1} \in \mathbb{R}$,
where $R_{t+1} = r$ is some numerical reward at timestep $t+1$. Finally, the
agent is in a new state $S_{t+1} = s'$.  As usual $\gamma$ is the discount factor.
Furthermore, we define a POMDP $(\mathcal{S}, \mathcal{A}, \mathcal{O}, \mathcal{P}, \mathcal{R}, \mathcal{Q},
\gamma)$ for egocentric 3-D environments, where the true state $s$ is not known. Thus, the agent sees the state $s \in \mathcal{S}$
through an observation $o \in \mathcal{O}$, where an observation function
$\mathcal{Q}: \mathcal{S} \rightarrow \mathcal{O}$ maps the true state to the
agent's observation. The components of our method are as follows: we define a policy network $\pi_\theta: \mathcal{O} \rightarrow \mathcal{A}$ with network parameters ${\theta}$, 
which transforms an observation $o$ into an action $a$. 

We utilize a pre-trained model $f_\nu: \mathcal{O} \rightarrow \mathcal{E}$ with network parameters $\nu$ to project observations into a 
latent space $\mathcal{E}$, which will be learned by self-supervised learning \citep[e.g. DINO, ][]{2021CaronDINO}. A clustering algorithm $g: \mathcal{E} \rightarrow \{1,\ldots M\}$ is also defined, 
which assigns a cluster label $m$ to an embedding $e$. 

Our method has two hyperparameters: 
the cosine similarity threshold $\kappa$ and the number of clusters $M$ for the episodic Gaussian mixture model. 
This is less than other exploration methods based on clustering.
In the following we describe our algorithm. 
We provide the Python implementation \footnote{We release our code on Github: \url{https://github.com/stefanwm13/just-cluster-it-public}}
for the following steps in Algorithm \ref{alg:methodalg}.

\begin{enumerate}
\item \textbf{Building episodic clusters.}
We define an episode as a sequence of transitions $\big\{(o^{(t)}, a^{(t)}, r^{(t)}, o'^{(t)})\big\}_{t=1}^{T}$, where $T$ is the length of the episode and $o'^{(t)}$ is the observation following $o^{(t)}$. 
Moreover, we define the set of observations as $B = \{o^{(t)}{\}}_{t=1}^{T}$ and 
let $\tilde{B} = \{e^{(t)}{\}}_{t=1}^{T} =  f_\nu(B)$ be the embeddings after passing them through the pre-trained model $f_\nu$.
We then fit a Gaussian mixture model (GMM) on the embeddings $\tilde{B}$ to obtain $M$ episodic cluster centers
\begin{equation} \{\mu_{\text{ep}}^{(m)}\}^M_{m=1}.\end{equation}
The GMM induces the mapping $g:\mathcal{E}\rightarrow\{1,\ldots M\}$ which assigns each embedding $e^{(t)}$ to one of the clusters, which defines a sequence of cluster labels $m_*(t)= g(e^{(t)})$.


\item \textbf{Updating the global cluster table.}
The following steps 2.1 to 2.4 are repeated for every episodic cluster $m=1, \ldots, M$.  This builds the global cluster table incrementally.

\begin{enumerate}
\item[2.1] \textbf{Assigning an episodic cluster center to global cluster centers.}
Before the first episode, the global cluster table is empty.  In step 2.3 new pairs will be added.
Assume for now that we have already $K$ pairs $(\mu_{\text{global}}^{(k)}, \lambda_{\text{global}}^{(k)})$ of global cluster centers and global cluster pseudo counts
\begin{equation}\big\{ (\mu_{\text{global}}^{(k)}, \lambda_{\text{global}}^{(k)}) \big\}_{k=1}^{K}.\end{equation}

Next, we calculate for fixed $m$ the pairwise cosine similarities between $\mu_{\text{ep}}^{(m)}$ and $\mu_{\text{global}}^{(k)}$, which are collected in a $K$-dimensional vector D. 
The maximum of $D$ gives us the maximal cosine similarity $d_*(m)$ and a corresponding global cluster label $k_*(m)\in\{1,\ldots,K\}$ for each episodic cluster center.

\item[2.2] \textbf{Calculating the intrinsic rewards.}
To calculate the intrinsic rewards along the episode, we assign the pseudo-counts for the embeddings $e^{(t)}$ that are assigned to cluster $m$,
\begin{equation}\rho^{(t)} = \lambda_\text{global}^{(k_*(m))} + \sum^{t}_{t'=1}{[m_*(t')=m]}\end{equation}
such that the pseudo-counts increase for each time an embedding belonging to the cluster $m$ is seen in the episode. 

For time steps $t$ where $\rho^{(t)}$ has just been assigned, we calculate the intrinsic rewards as the inverse square root of the pseudo-counts \citep{Strehl2008Counts},
\begin{equation}r_{\text{int}}^{(t)} =  \frac{1}{{\sqrt{\rho^{(t)}}}}.
\end{equation}

\item[2.3]\textbf{Growing the global cluster table.}  If the maximal cosine similarity $d_*(m)$ is smaller than the hyperparameter $\kappa$ (a fixed value between 0 and 1), we add $\mu_\text{ep}^{(m)}$ to the list of global cluster centers and $K$ is increased by one.  The corresponding global pseudo count is initialized to zero, but will get updated in the next step.
 
\item[2.4] \textbf{Updating the global pseudo counts.}
The global pseudo count of the global cluster center that corresponds to $m$, i.e., with index $k_*(m)$, is simply updated using the episodic cluster counts for the assigned episodic cluster center $m$:
\begin{equation}
    \lambda_\text{global}^{(k_*(m))} := \lambda_\text{global}^{(k_*(m))} + \lambda_\text{ep}^{(m)}.
\end{equation}
\end{enumerate}
\item \textbf{Combining intrinsic and extrinsic rewards.}  Finally, the intrinsic and extrinsic rewards are simply added, and we update the policy network $\pi_\theta$ with the transitions 
\begin{equation}\{(o^{(t)}, a^{(t)}, r_{\text{int}}^{(t)} +  r^{(t)} , o'^{(t))}{\}}_{t=1}^{T}\end{equation}
via Proximal Policy Optimization (PPO).
\end{enumerate}

\begin{figure*}[htbp]
	\centering 
	\begin{tikzpicture}
		\node[draw, thin, color=white, fill=white, rounded corners=0.0mm, inner sep=1mm] (img1) {
			\includegraphics[scale=0.32]{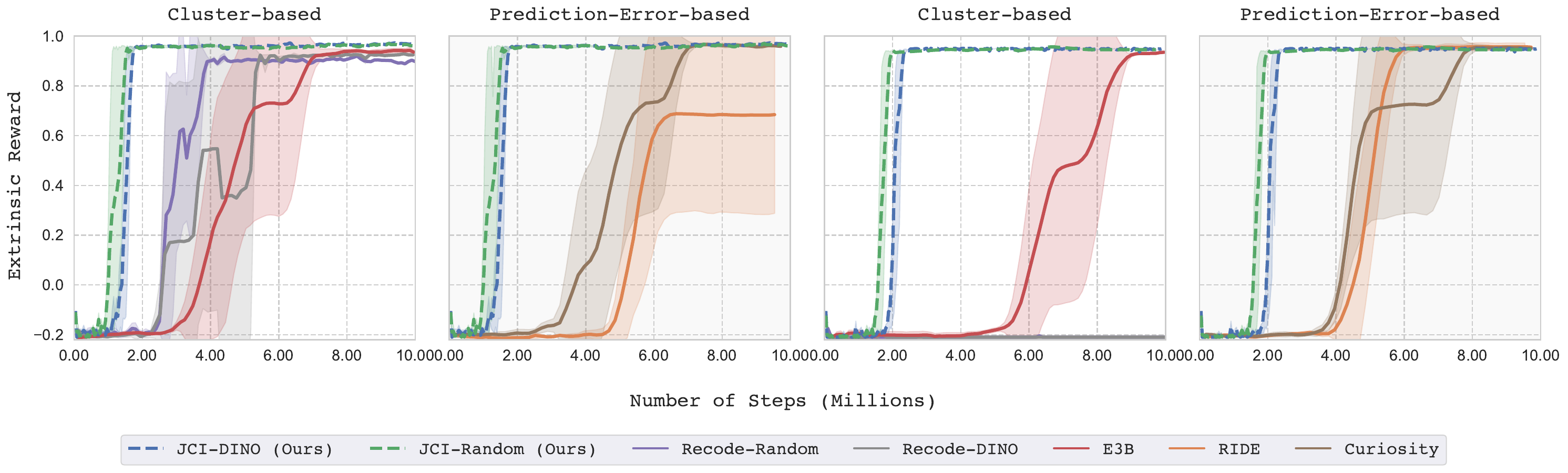}
		};
		\node[text width=6cm, text centered, above of=img1] at (-3.5, 1.7) {\fontfamily{pcr}\selectfont\scriptsize{MyWayHome - Sparse}};	
		\node[text width=6cm, text centered, above of=img1] at (3.9, 1.7) {\fontfamily{pcr}\selectfont\scriptsize{MyWayHome - VerySparse}};
	\end{tikzpicture}
	\caption{\textbf{Clustering random features is effective for VizDoom:} Our method with
		clustering outperforms other traditional exploration methods.
		Interestingly, random features perform slightly better than the
		DINO features for VizDoom. While the complexity of the observations is not trivial, in this setting 
		the priors in the pre-trained representations seem to be not more informative than random features. This also shows that 
		in 3-D high-dimensional environments random features are salient enough when clustered to estimate pseudo-counts.}
		\label{fig:vizdoom_results}
\end{figure*}

\section{Experiments}
\label{sec:experiments}
Given our proposed ideas and method we aim to answer the following questions: (i) How well does clustering features perform in high-dimensional 3-D environments?
(ii) Can our method leverage pre-trained representations to improve exploration? 
(iii) What effect do episodic clustering and global clustering have on the performance and robustness of our method?
We present the environments, experiment setup and baselines, followed by results of our experiments and corresponding ablations.

\paragraph{Environments.}
We test our method on observationally complex environments like VizDoom \citep{Kempka2016ViZDoom} and Habitat-Replica \citep{habitat19iccv,szot2021habitat,puig2023habitat3}.
In VizDoom every room has a different texture, where some moving textures are also present. In Habitat,
the observations are real world scenes of apartments, hotels and office rooms which are very complex. 
The chosen environments are centered around navigation exploration tasks. This setup enables us to separate the 
issue of density estimation from the challenge of understanding complex environment dynamics. 
In general, all methods we evaluate including ours focus solely on the problem 
of density estimation, and we thus share the same assumptions. More details regarding the environments and agent setup can be found in the Appendix \ref{sec:environments}.

\begin{figure*}[htbp]
	\centering
	\begin{tikzpicture}
		\node[draw, thin, color=white, fill=white, rounded corners=0.0mm, inner sep=1mm] (img1) {
			\includegraphics[scale=0.35]{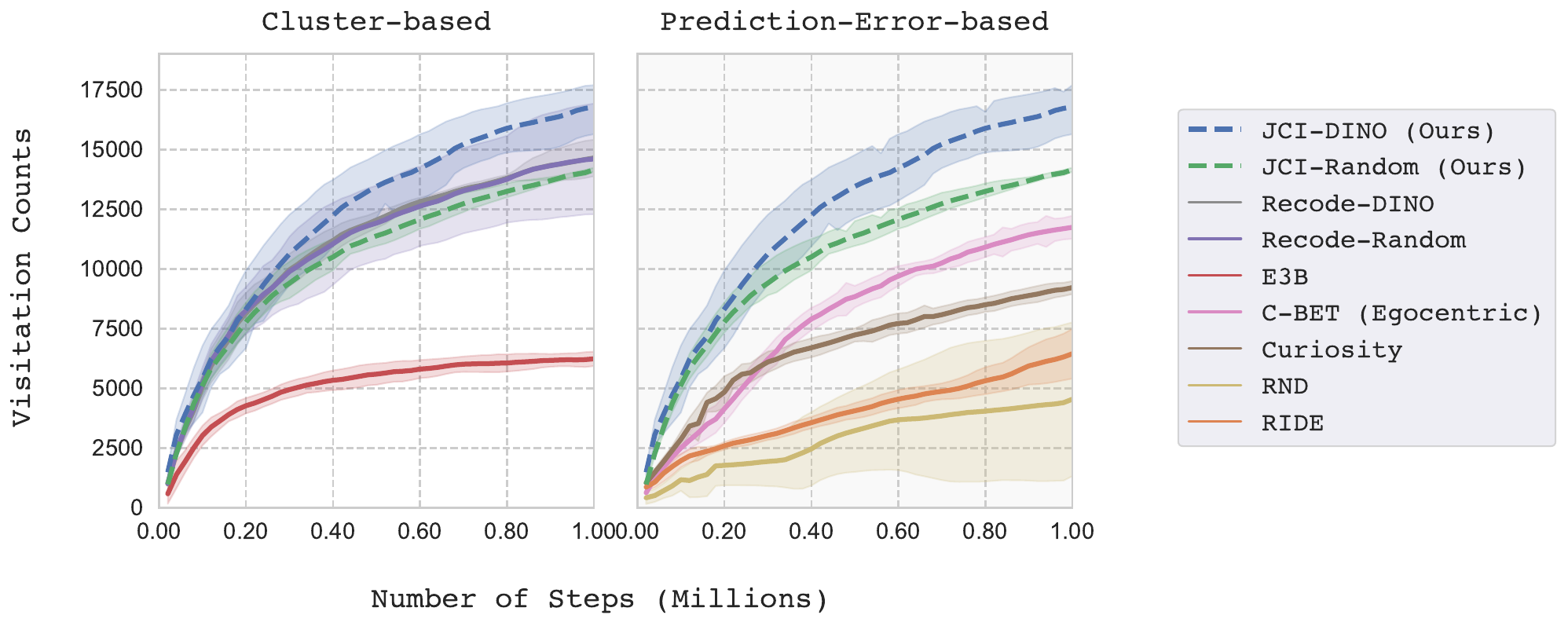}
		};
		\node[text width=6cm, text centered, above of=img1] at (-1.4, 1.7) {\fontfamily{pcr}\selectfont\scriptsize{Habitat - Apartment-0}};	
	\end{tikzpicture}

	\caption{\textbf{Pre-trained DINO features excel with more complex observations:} When training on the Habitat environment we see that clustering with DINO embeddings 
	is more effective than clustering with random features. Furthermore, only our method is able to leverage the pre-trained DINO features effectively. 
	We argue that the priors present in the DINO embeddings help in agreggating the representations when building the episodic clusters, which ultimately determine the pseudo-counts.} 
	\label{fig:habitat_results}
\end{figure*}

\paragraph{Experimental setup and baselines.}
We compare our method to both clustering and prediction-error-based exploration approaches. Thus, we split the experiments into two categories for both environments. 
Both Figures \ref{fig:vizdoom_results} and \ref{fig:habitat_results} show comparison to only (i) clustering-based methods on the left side
 and to (ii) prediction-error-based methods on the right side.
 For (i) we compare against RECODE \citep{2023SaadeUnlocking} and E3B \citep{2022HenaffE3B}. For (ii) we compare against 
Random Network Distillation \citep{2018BurdaRandom}, Curiosity-driven exploration \citep{2017PathakCuriosity} and RIDE \citep{2020RaileanuRIDE}. 
To see how other methods leverage pre-trained representations, we also compare against RECODE with both random features and DINO features.
For all configurations we perform 3 runs with different seeds. We use PPO \citep{2017SchulmanPPO} to train the policy network for both our method and our implementation of RECODE. 
E3B \citep{2023HenaffExploration} also uses DD-PPO for the Habitat environment. For all other methods the learning method for the policy network is IMPALA \citep{2018EspeholtIMPALA}. 
Specific implementation details and hyperparameters can be found in Appendix \ref{sec:implementation_details}.

\paragraph{Evaluation metrics.} For VizDoom we use the extrinsic reward as a metric for performance. The extrinsic reward is defined from $0$ to $1$. 
For Habitat, we explore only using the intrinsic reward, we thus use the number of visited states as a metric for performance as in \cite{parisi2021interesting}.
We show the exact formula in Appendix \ref{sec:habitat_env}.	


\subsection{VizDoom Results---The Suprising Effectiveness of Random Features}
We plot results for  "MyWayHome" in Figure \ref{fig:vizdoom_results}, where we compare our method to both cluster-based (left) and prediction-error-based methods (right) for both
the ``Sparse'' and ``VerySparse'' settings.
We see that clustering random features with our method (dashed, green curve) is superior to all other methods converging at around 1.5M and 1.8M steps respectively. 
Interestingly, random features perform slightly better in VizDoom than the
DINO features (dashed, blue curve) which converges slightly later at around 1.7M and 1.9M steps respectively. 
In this case, the priors in the DINO features do not seem to pose an advantage over random priors for the observations in the VizDoom environment. However, this also shows 
that random features in 3-D environments can be salient enough to be clustered effectively.
When comparing our method to RECODE with DINO features (grey curve), we see that our method is better able to leverage the pre-trained
features. For RECODE, DINO features converge at around 5M steps compared to the random features (purple curve) which converge at 3.8M steps, 
while for our method random features and DINO features converge at almost the same time.
For ``VerySparse'' we could not get RECODE to converge within the specified number of steps. One explanation for this could be 
that in the ``VerySparse'' setting, the agent never reaches the goal by chance as opposed to in the ``Sparse'' setting, such that the agent never finds a sensible policy. 
The RECODE agent does however manage to reach the goal on a few occasions.
Furthermore, comparing our method to the prediction-error-based methods such as RIDE and Curiosity, we see that these methods perform worse needing at least 5M steps to converge. 
Overall, both our clustering method and RECODE, outperform the prediction-error-based methods. While E3B is a clustering method as well, it only performs episodic clustering as explained in Section \ref{sec:related_work}.  
Usually, episodic clustering alone performs worse in singleton environments, such as VizDoom as explained in \citet{2023HenaffExploration}.

\subsection{Habitat Results---Pre-trained features help in complex observations}
\begin{figure*}
    \centering
    \subfigure[]{ 		
		\begin{tikzpicture}
			\node[draw, thin, color=white, fill=white, rounded corners=0.0mm, inner sep=1mm] (img1) {
				\includegraphics[width=0.33\textwidth]{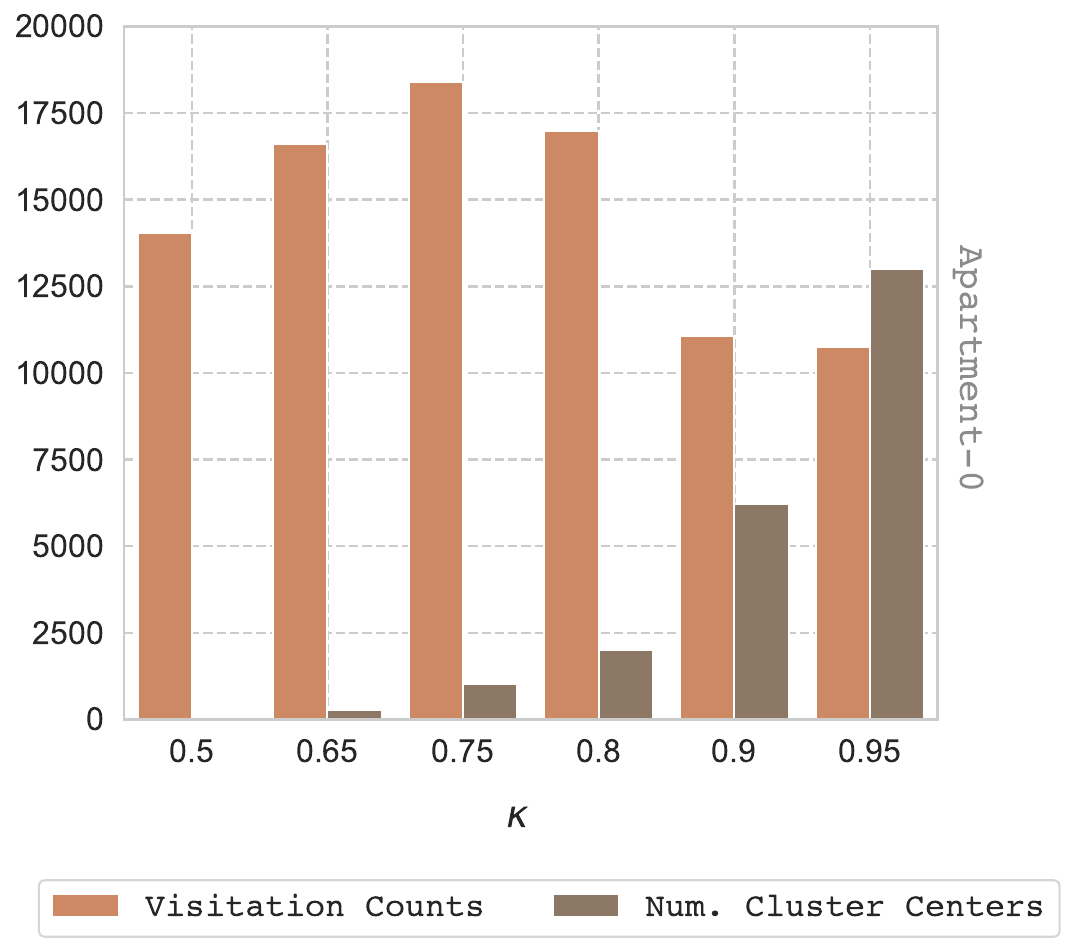}
			};
			\node[text width=6cm, text centered, above of=img1] at (0, 1.7) {\fontfamily{pcr}\selectfont\scriptsize{Ablation - Cosine Similarity Threshold $\kappa$}};	
		\end{tikzpicture}
		\label{fig:bandwidth}
	}\hfil
	\subfigure[]{
		\begin{tikzpicture}
			\node[draw, thin, color=white, fill=white, rounded corners=0.0mm, inner sep=1mm] (img2){
				\includegraphics[width=0.55\textwidth]{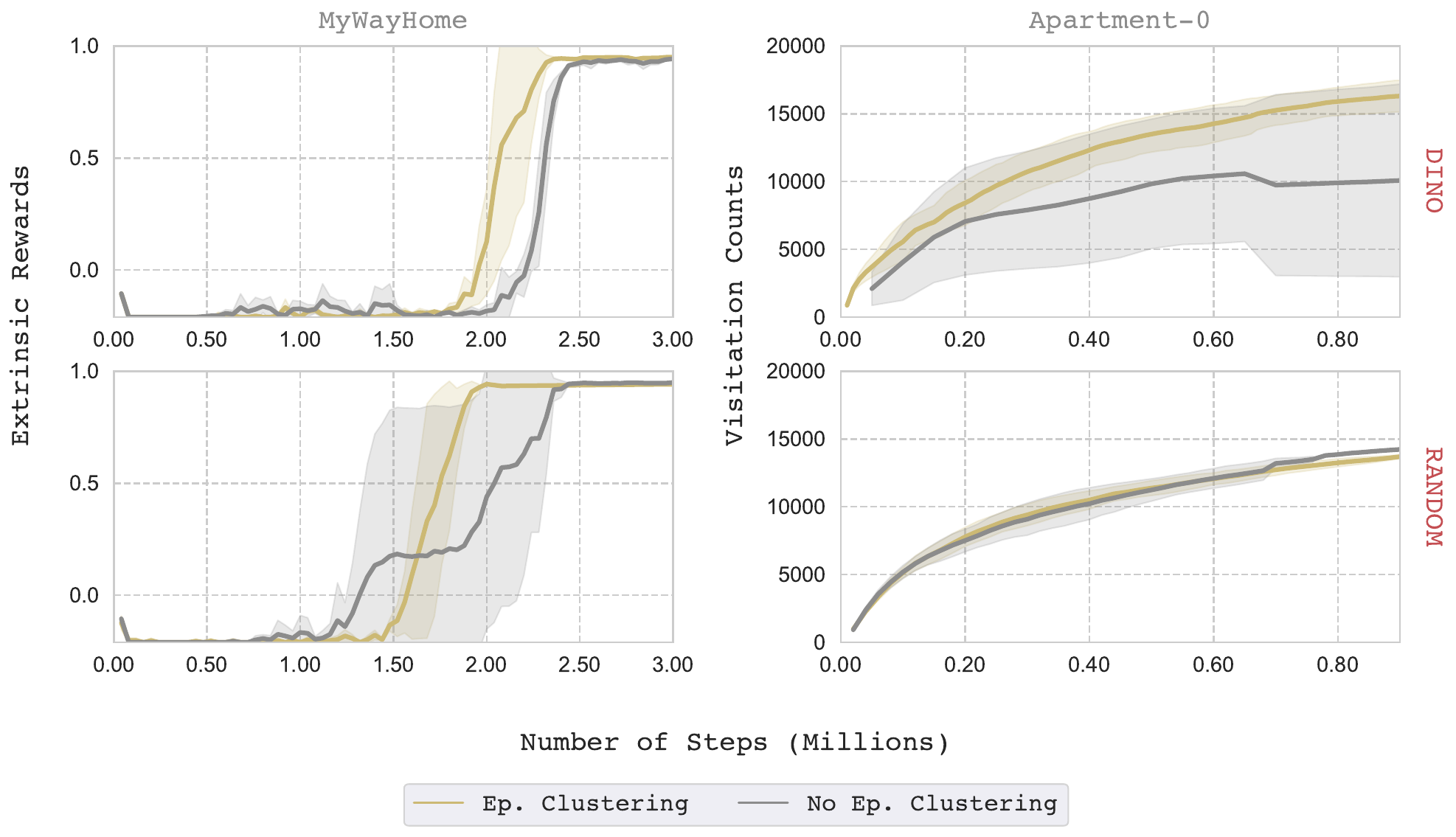}
			};
			\node[text width=6cm, text centered, above of=img2] at (0, 1.9) {\fontfamily{pcr}\selectfont\scriptsize{Ablation - Episodic Clustering}};	
		\end{tikzpicture}
		\label{fig:episodic_cluster}

	} 
    \caption{\textbf{Ablation studies: We show the effectiveness of clustering, especially episodic clustering:} \textbf{(a)} Increasing the cosine similarity threshold $\kappa$ increases the number of global clusters in the cluster table. 
	If the number of clusters in the cluster table are too granular, performance decreases as shown by lower visitation counts for higher values of $\kappa$.
	 \textbf{(b)} Episodic clustering helps whenever there is structure in the representations. Especially for DINO in Habitat performance drops significantly, if episodic clustering is removed.
	 This can also be seen in VizDoom (left panels) where episodic clustering helps both DINO and random features. For Habitat (right panels) with random features, episodic clustering has no large effect
	 since the random features are not expressive enough for the complex observations.}
\end{figure*}

To test our method on more complex observations we use the Habitat environment and adopt the experimental setting from \citet{parisi2021interesting}. 
That is, we train the agent on the Habitat environment with only intrinsic rewards and plot the visitation counts (see Figure \ref{fig:habitat_results}).
We can see the benefit of using pre-trained DINO features in the visually more complex Habitat environment.
When clustering pre-trained DINO features with our method (dashed, blue curve), we achieve 17K visitation counts on average, more than any other method. RECODE by 
comparison achieves only slightly under 15K visitation counts. Interestingly, our method with random features (dashed, green curve) performs slightly worse than RECODE at 14K visitation counts.
We show in Figure \ref{fig:episodic_cluster} that for Habitat, episodic clustering of random features is not better than episodic clustering of DINO features, hence the reduced performance compared to DINO features.
However, we see again that our method is able to better leverage the pre-trained features compared to RECODE. Again, E3B is not suited for singleton environments barely achieving 7K visitation counts. 
Looking at the prediction-error-based methods, although Random Network Distillation uses random 
features as well, it is vastly inferior to the clustering-based methods converging at around 5K visitation counts. 
This validates our hypothesis about \emph{transition saliency} presented in Section \ref{sec:introduction}:
In 3-D environments, the clustering-based methods are able to exploit the structure of the representations more effectively. We will 
expand on this in the following section.

\subsection{The Effect of Episodic and Global Clustering}
Next, we analyse the effect of the episodic clustering and the global clustering. 
To ablate the episodic clustering, we perform runs for both "MyWayHome-VerySparse" and "Apartment-0" with and without episodic clustering. To 
ablate the influence of the global clustering, we vary the cosine similarity threshold $\kappa$ and plot the number of cluster centers and visitation counts for ``Apartment-0''.

\paragraph{Increasing cosine similarity threshold $\kappa$ makes pseudo-counts more granular, degrading performance.} We plot the visitation counts and the number of clusters in the cluster table for
different values of $\kappa$ in Figure \ref{fig:bandwidth}.
We see that increasing the cosine similarity threshold $\kappa$ increases the number of global cluster centers in the cluster table.
If the number of clusters in the cluster table is too high, performance decreases as shown by lower visitation counts for higher values of $\kappa$.
We argue that the pseudo-counts become too sparse, and thus the intrinsic reward fails to provide a good approximation of the state-space distribution.
This aligns with the \emph{transition saliency} problem in Section \ref{sec:introduction}. In this case we revert to the same setting of prediction-error-based methods where the intrinsic reward becomes too granular.
Note that the performance at around 10K visitation counts is still better than random network distillation \citep{2018BurdaRandom} in Figure \ref{fig:habitat_results},
which is likely related to the fact that clustering-based methods are not affected by catastrophic forgetting.

Interestingly, for lower values of $\kappa$ performance does not degrade in the same way. 
We observe that the smaller the value of $\kappa$ is the less cluster centers get added to the cluster table. 
Hence the intrinsic reward is defined over fewer cluster centers. We see in Figure \ref{fig:bandwidth} that up to $\kappa=0.5$ performance remains comparable. 

\paragraph{Episodic clustering is effective with good representations.} We plot visitation counts and extrinsic reward with and without episodic clustering
in Figure \ref{fig:episodic_cluster}.
We see that episodic clustering helps whenever the underlying representations are structured. This can be seen for Habitat (right panels) with DINO features where performance drops
dramatically without episodic clustering (upper right). Even for VizDoom (left panels), episodic clustering helps both with DINO and random features, which shows that
there is hidden structure to exploit in the comparatively simpler observations. For Habitat with random features, 
episodic clustering has no big effect since the random features are probably not expressive enough for the complex observations. We see this in 
Figure \ref{fig:comparison} (in the Appendix) where the random features are not clustered effectively compared to the DINO features.
Overall, during our tests we found that 10\% of the total number of steps in an episode is a good number of clusters for the Gaussian mixture model.

\begin{figure*}[htbp!]
    \centering
		\begin{tikzpicture}
			\node[draw, inner sep=0.5mm, thin, rounded corners=1mm] at (0,0) {\includegraphics[width=.155\textwidth]{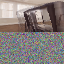}};
			\node[draw, inner sep=0.5mm, thin, rounded corners=1mm] at (2.8,0) {\includegraphics[width=.155\textwidth]{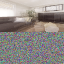}};
			\node[draw, inner sep=0.5mm, thin, rounded corners=1mm] at (5.6,0) {\includegraphics[width=.155\textwidth]{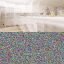}};
			\node[draw, inner sep=0.5mm, thin, rounded corners=1mm] at (8.4,0) {\includegraphics[width=.155\textwidth]{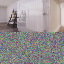}};
			\node[draw, inner sep=0.5mm, thin, rounded corners=1mm] at (11.2,0) {\includegraphics[width=.155\textwidth]{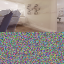}};
			\node[draw, inner sep=0.5mm, thin, rounded corners=1mm] at (14,0) {\includegraphics[width=.155\textwidth]{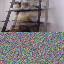}};		
		\end{tikzpicture}
	
	\caption{\textbf{Visualization of DINO clusters for observations with random noise:} We plot different samples from the cluster table for the noisy observations in Habitat. Each image represents the mean image of an episodic cluster that was aggregated to the cluster table. The clustering happens on the 384-dimensional embedding. Even with the concatenated random noise the embeddings are still clustered sensibly.} 
    \label{fig:vizdoom_cluster_vis_noise}
\end{figure*}

\subsection{Clustering is Robust to Environmental Noise}
\begin{figure}[htbp!]
    \centering
		\begin{tikzpicture}
			\node[draw, thin, color=white, fill=white, rounded corners=0.0mm, inner sep=1mm] (img1) {
				\includegraphics[width=0.16\textwidth]{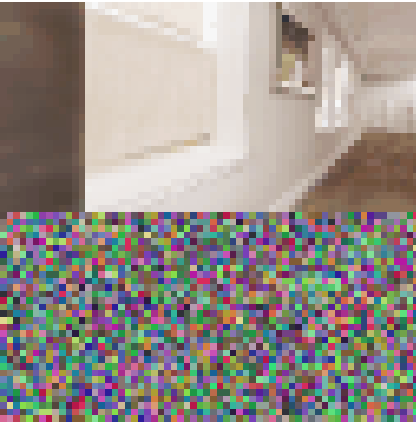}
			};
		   \node[text width=6cm, text centered, above of=img1] at (0, 0.7) {\fontfamily{pcr}\selectfont\scriptsize{Observation with Random Noise}};	

		\end{tikzpicture}
    \caption{\textbf{We concatenate random noise to the observations:} To test the robustness of our method to observations with random noise, i.e, the noisy TV problem, we perform the same experiment as \citet{2023SaadeUnlocking} and concatenate a frame with random noise that changes after every step to the observation. }
    \label{fig:noise_obs}
\end{figure}

To the test the effect of random noise on exploration like in the noisy-TV problem \citep{2017PathakCuriosity}, we adopt the experimental setup of \citet{2023SaadeUnlocking}, where a frame of random noise is concatenated to the current observation. 
Note that for each step the different frame with random noise is sampled. We show an example of the observation with the concatenated noise in Figure \ref{fig:noise_obs}.

\begin{figure}[htbp!]
    \centering
		\begin{tikzpicture}
			\node[draw, thin, color=white, fill=white, rounded corners=0.0mm, inner sep=1mm] (img1) {
				\includegraphics[width=0.4\textwidth]{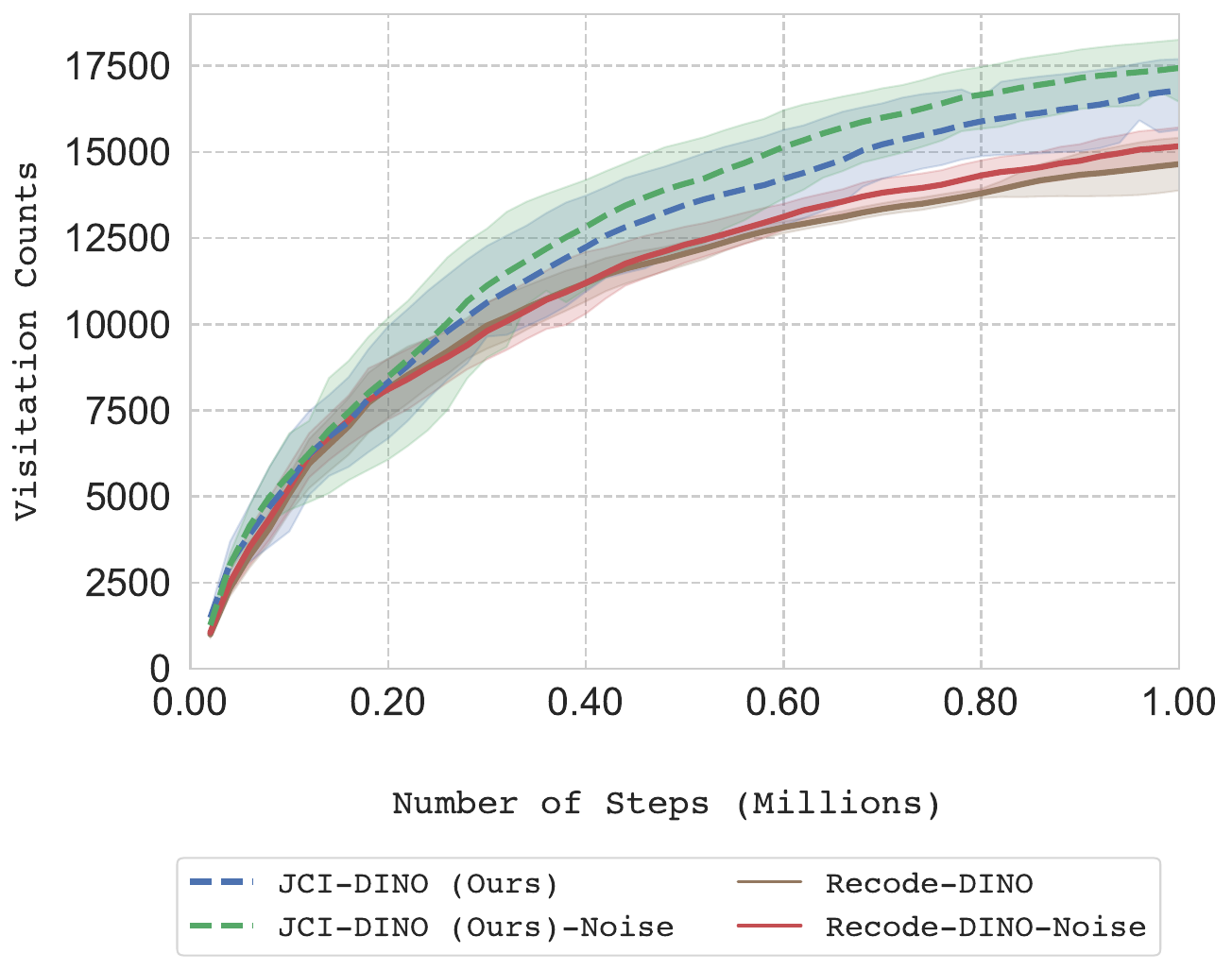}
			};
			\node[text width=8cm, text centered, above of=img1] at (0, 2) {\fontfamily{pcr}\selectfont\scriptsize{Apartment-0 - Exploration with noisy observations}};	
		\end{tikzpicture}
		\caption{\textbf{Our method is robust to random noise in the observations:}  We see that our method is robust to random noise in the observations. As a sanity check we also evaluate RECODE on noisy observations and confirm the results from \citet{2023SaadeUnlocking}. Suprisingly, for both methods 
		the performance increases when adding random noise. We suspect the random noise has a regularising effect on the clustering.}    
		\label{fig:noise_results}
\end{figure}

\paragraph*{Random noise does not affect exploration for clustering based methods.} We rerun the Habitat Apartment-0 with noisy observations for both our method and RECODE in Figure \ref{fig:noise_results}. We show that our method is unaffected by random noise in the observations. Interestingly, performance slightly increases with noisy observations for both our method and RECODE. When looking at the number of cluster entries, we observe that when using observations with random noise the amount of clusters in the cluster table is lower than without the random noise. We therefore suspect that the random noise has a regularising effect on the clustering.

We also show the saved cluster centers in the cluster table with noisy observations in Figure \ref{fig:vizdoom_cluster_vis_noise}. The quality of the clustering seems unaffected since similar observations are still aggregated correctly despite the random noise.

\section{Conclusion}
The starting point of our work is the insight that the magnitude of pixel-changes can be large in 3-D environments, but usually lacks in saliency.
This is problematic in exploration with sparse rewards, since the agent is not able to distinguish between important and unimportant changes in the observations.
To mitigate this problem, we proposed a method for state aggregation based on episodic and global
clustering of pre-trained representations.  We have shown that pre-trained representations and also random features can be leveraged effectively 
in environments with complex observations. We ablated different components of our method:  we show that 
the cosine similarity threshold $\kappa$ is vital to control the granularity of the clusters in the cluster table.   We notice that
performance degenerates if the number of clusters in the cluster table is too high, since the pseudo-counts become too sparse and the 
intrinsic reward too noisy. 
Furthermore, we found out that episodic clustering 
leverages the priors in the representations in order to produce better pseudo-count estimates, 
thus playing a crucial role for complex observations since the counts for the corresponding representations are aggregated more effectively.

\paragraph{Limitations and future work.}
So far, we tested our method on singleton MDP's. We believe our method can also be extended to contextual MDPs, possibly by counting states separately on an episodic level. 
Overall, our method sheds light onto the problem of transition saliency in 3-D environments which is ubiquitous in exploration with sparse-rewards.
In this sense, an interesting direction for future work could be investigating whether mitigating the transition saliency problem can enhance the effectiveness of 
prediction-error-based methods in 3-D enviroments. Hypothetically, clustering representations and using them as fixed targets for prediction-error-based methods should result in a more salient intrinsic reward. 
Future research should continue investigating which kinds of representations contain the best priors for exploration with sparse rewards. 
Potentially, multi-modal generative models could be used to learn representations that not only capture visual invariances, but also environmental dynamics. 

\section*{Impact Statement}
This paper presents work whose goal is to advance the field of Machine Learning. There are many potential societal consequences of our work, none which we feel must be specifically highlighted here.

\nocite{*}

\bibliography{example_paper}
\bibliographystyle{icml2024}
\newpage
\appendix
\onecolumn
\section{Algorithm}
\label{sec:algorithm}

\begin{lstlisting}[language=Python, caption=Just Cluster It! Algorithm, label=alg:methodalg]
	# Hyperparameters
	kappa = 0.8 # Cosine similarity threshold
	M = 250 # Number of episodic clusters

	# Global variables
	self.gmm = GaussianMixture(n_components=M, reg_covar=1e-6, covariance_type="full", random_state=1, warm_start=False)
	self.cluster_table = [np.zeros(embedding_size), 0] # (Equation 2)

	def update_cluster_table(self, embeddings, rewards):		
		# Step 1 (Build episodic clusters) 
		self.gmm.fit(embeddings) # Fit GMM (Equation 1)
		m_prime = self.gmm.predict(embeddings) # Get cluster labels

		# Assign embeddings to episodic clusters
		cluster_dict = {}
		for index, embedding in enumerate(embeddings):
			if m_prime[index] not in cluster_dict:
				cluster_dict[m_prime[index]] = []
			cluster_dict[m_prime[index]].append([embedding])

		# Step 2 (Updating the global cluster table)
		for m in cluster_dict: 
			episodic_cluster = np.stack(cluster_dict[m])
			lambda_ep = episodic_cluster.shape[0]

			# Calculate episodic cluster center 
			mu_ep = np.mean(np.stack(episodic_cluster[:, 0]), 0) # (Equation 1)
			mu_globals = np.stack(self.cluster_table[:, 0])

			# Step 2.1 (Assigning episodic cluster center(s) to global cluster centers)
			distances = cosine_similarity(np.expand_dims(mu_ep, 0), mu_globals)
			k_prime = np.argmax(distances)
			d_prime = np.max(distances)

			# Step 2.2 (Calculating intrinsic rewards)				
			iverson_bracket_count = 0
			max_lambda_global = self.cluster_table[k_prime, 1]
			lambda_global = 0 if d_prime < kappa else max_lambda_global

			for index, _ in enumerate(embeddings): 
				# (Equation 3)
				if m_prime[index] == m:
					iverson_bracket_count += 1 
					rho = lambda_global + iverson_bracket_count 

					# (Equation 4)
					ir = 1 / np.sqrt(rho) 
					rewards[index] += 0.1*ir # Scale intrinsic reward with 0.1

			# Step 2.3 (Growing the cluster table)
			if d_prime < kappa: 
				self.cluster_table = np.vstack((self.cluster_table, [mu_ep, lambda_ep])) 
			else:
				# Step 2.4 (Updating the global pseudo counts)
				self.cluster_table[k_prime, 1] += lambda_ep  # (Equation 5)

		return rewards # Step 3


	\end{lstlisting}

\section{Visualizations}
\subsection{What do the clusters look like?}
\FloatBarrier
\begin{figure}[htbp!]
    \centering
	\subfigure{
		\begin{tikzpicture}
			\node[draw, inner sep=0.5mm, thin, rounded corners=1mm] at (0,0) {\includegraphics[width=.155\textwidth]{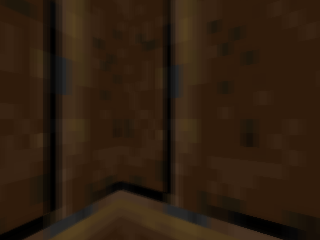}};
			\node[draw, inner sep=0.5mm, thin, rounded corners=1mm] at (2.8,0) {\includegraphics[width=.155\textwidth]{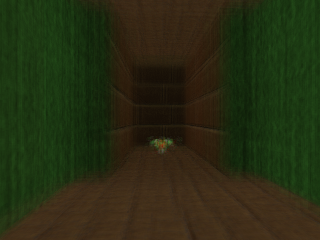}};
			\node[draw, inner sep=0.5mm, thin, rounded corners=1mm] at (5.6,0) {\includegraphics[width=.155\textwidth]{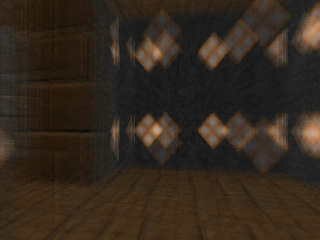}};
			\node[draw, inner sep=0.5mm, thin, rounded corners=1mm] at (8.4,0) {\includegraphics[width=.155\textwidth]{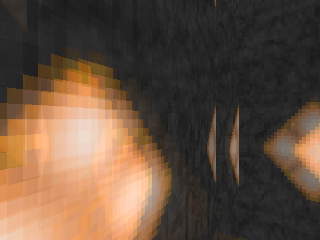}};
			\node[draw, inner sep=0.5mm, thin, rounded corners=1mm] at (11.2,0) {\includegraphics[width=.155\textwidth]{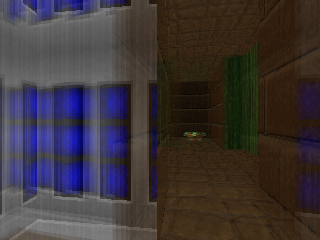}};
			\node[draw, inner sep=0.5mm, thin, rounded corners=1mm] at (14,0) {\includegraphics[width=.155\textwidth]{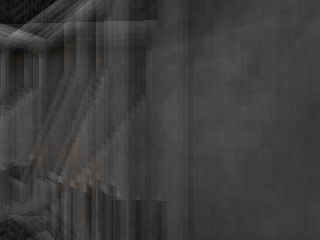}};
			
			\node[draw, inner sep=0.5mm, thin, rounded corners=1mm] at (0,-3) {\includegraphics[width=.15\textwidth]{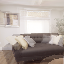}};
			\node[draw, inner sep=0.5mm, thin, rounded corners=1mm] at (2.8,-3) {\includegraphics[width=.15\textwidth]{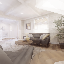}};
			\node[draw, inner sep=0.5mm, thin, rounded corners=1mm] at (5.6,-3) {\includegraphics[width=.15\textwidth]{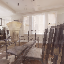}};
			\node[draw, inner sep=0.5mm, thin, rounded corners=1mm] at (8.4,-3) {\includegraphics[width=.15\textwidth]{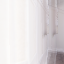}};
			\node[draw, inner sep=0.5mm, thin, rounded corners=1mm] at (11.2,-3) {\includegraphics[width=.15\textwidth]{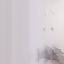}};
			\node[draw, inner sep=0.5mm, thin, rounded corners=1mm] at (14,-3) {\includegraphics[width=.15\textwidth]{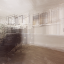}};
				
		\end{tikzpicture}
	}
	\caption{\textbf{Visualization of DINO clusters for VizDoom and Habitat:} We plot different samples from the cluster table for VizDoom. Each image represents the mean image of an episodic cluster that was aggregated to the cluster table. 
	The clustering happens on the 384-dimensional embedding. We can see that observations are aggregated sensibly and account for similar states, where having a separate pseudo-count would not be beneficial.
	By looking at the counts and appearances we can also see that the clusters are not too granular, since the counts are not too low and the same cluster center is encountered multiple times across different episodes.} 
\label{fig:vizdoom_cluster_vis}
\end{figure}
\FloatBarrier

We visualize some of the clusters centers stored in the global cluster table, which are technically episodic clusters centers that were aggregated to the global cluster table.
We show the mean image of each cluster center in Figure \ref{fig:vizdoom_cluster_vis} for VizDoom and for Habitat. Note that the actual
clustering was performed with the 384-dimensional DINO embedding, but we show the images for better interpretability. We can see that the clusters are aggregated sensibly and account for similar states, 
where having a separate pseudo-count would not be beneficial. Especially for Habitat, the complex observations are aggregated to similar states, which is a testament to the effectiveness of the DINO features.

\subsection{Clustering DINO vs Random Features}
\begin{figure}[htbp!]
    \centering
	\subfigure{
		\begin{tikzpicture}
			\node[draw=magenta, inner sep=0.5mm, thick, rounded corners=1mm] (img1) at (0,0) {\includegraphics[width=.13\textwidth]{figures/habitat_cluster/high/img4_53_9.png}};
			\node[draw=magenta, inner sep=0.5mm, thick, rounded corners=1mm] (img2) at (2.6,0) {\includegraphics[width=.13\textwidth]{figures/habitat_cluster/high/img5_620_60.png}};
			\node[draw=magenta, inner sep=0.5mm, thick, rounded corners=1mm] (img3) at (5.2,0) {\includegraphics[width=.13\textwidth]{figures/habitat_cluster/high/img6_814_84.png}};
			\node[draw=magenta, inner sep=0.5mm, thick, rounded corners=1mm] (img4) at (7.8,0) {\includegraphics[width=.13\textwidth]{figures/habitat_cluster/high/img7_5551_701.png}};
			\node[draw=magenta, inner sep=0.5mm, thick, rounded corners=1mm] (img5) at (10.4,0) {\includegraphics[width=.13\textwidth]{figures/habitat_cluster/high/img8_1199_186.png}};
			\node[draw=magenta, inner sep=0.5mm, thick, rounded corners=1mm] (img6) at (13,0) {\includegraphics[width=.13\textwidth]{figures/habitat_cluster/high/img9_1366_781.png}};

			\node[draw=cyan, inner sep=0.5mm, thick, rounded corners=1mm] at (0,4.5) {\includegraphics[width=.13\textwidth]{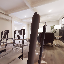}};
			\node[draw=cyan, inner sep=0.5mm, thick, rounded corners=1mm] at (2.6,4.5) {\includegraphics[width=.13\textwidth]{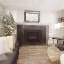}};
			\node[draw=cyan, inner sep=0.5mm, thick, rounded corners=1mm] at (5.2,4.5) {\includegraphics[width=.13\textwidth]{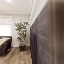}};
			\node[draw=cyan, inner sep=0.5mm, thick, rounded corners=1mm] at (7.8,4.5) {\includegraphics[width=.13\textwidth]{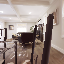}};
			\node[draw=cyan, inner sep=0.5mm, thick, rounded corners=1mm] at (10.4,4.5) {\includegraphics[width=.13\textwidth]{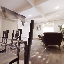}};
			\node[draw=cyan, inner sep=0.5mm, thick, rounded corners=1mm] at (13,4.5) {\includegraphics[width=.13\textwidth]{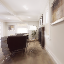}};

			\node[text width=4cm, text centered, above of=img1] at (0, 1.4) {Counts = 267 \\ Apps = 60};
			\node[text width=4cm, text centered, above of=img2] at (2.6, 1.4) {Counts = 686 \\ Apps = 128};
			\node[text width=4cm, text centered, above of=img2] at (5.2, 1.4) {Counts = 31481 \\ Apps = 4014};
			\node[text width=4cm, text centered, above of=img2] at (7.8, 1.4) {Counts = 1949 \\ Apps = 333};
			\node[text width=4cm, text centered, above of=img2] at (10.4, 1.4) {Counts = 2992 \\ Apps = 483};
			\node[text width=4cm, text centered, above of=img2] at (13, 1.4) {Counts = 27886 \\ Apps = 4909};

			\node[text width=4cm, text centered, above of=img1] at (0, -2.9) {Counts = 53 \\ Apps = 9};
			\node[text width=4cm, text centered, above of=img2] at (2.6, -2.9) {Counts = 620 \\ Apps = 60};
			\node[text width=4cm, text centered, above of=img2] at (5.2, -2.9) {Counts = 814 \\ Apps = 84};
			\node[text width=4cm, text centered, above of=img2] at (7.8, -2.9) {Counts = 5551 \\ Apps = 701};
			\node[text width=4cm, text centered, above of=img2] at (10.4, -2.9) {Counts = 1199 \\ Apps = 186};
			\node[text width=4cm, text centered, above of=img2] at (13, -2.9) {Counts = 1366 \\ Apps = 781};
		\end{tikzpicture}
	}
	\caption{\textbf{Comparison of \textcolor{magenta}{DINO clusters} and \textcolor{cyan}{Random clusters}:} From the observations in the Habitat environment we plot different samples from the cluster table for DINO (magenta) and random features (cyan).
	The episodic clusters from DINO and random features are fundamentally different. While the DINO clusters aggregate more dissimilar observations that share common features, the random clusters are more granular seen by the fact that the cluster mean images are free from distortions.
	From the counts we can see that random features still are able to aggregate counts, however very similar observations may not be aggregated as well as with the DINO clusters.} 
\label{fig:comparison}
\end{figure}
We compare the clusters from DINO (magenta) and random features (cyan) in Figure \ref{fig:comparison}. 
 We can see that the episodic clusters from DINO and random features are fundamentally different. Overall,
 the DINO cluster centers are much less granular compared to the random features. 
 This can be seen by the fact that the cluster mean images for the random features are free from distortions.
 Moreover, some cluster centers for the random features are very similar to each other, which is harder to find for the DINO features.
 From the counts we can see that random features still are able to aggregate counts, however very similar observations may not be aggregated as well as with the DINO clusters.
 This can be seen by the fact that some counts are unusually high for the random features indicating that some episodic clusters may be assigned erroneously to the same global cluster center.
 We were not able to find such high counts for the DINO features. Overall, we see that episodic clustering together with global clustering is effective 
 for both DINO and random features, since episodic cluster centers are matched multiple times throughout learning to the global cluster centers, as can be
 seen by the number of appearances ``Apps'' in Figure \ref{fig:comparison}.

 \newpage
 \section{Further Ablations}
 \subsection{Comparing DINO model size}
 \FloatBarrier
 \begin{figure*}[htbp!]
	 \centering
		 \begin{tikzpicture}
			 \node[draw, thin, color=white, fill=white, rounded corners=0.0mm, inner sep=1mm] (img2){
				 \includegraphics[width=0.53\textwidth]{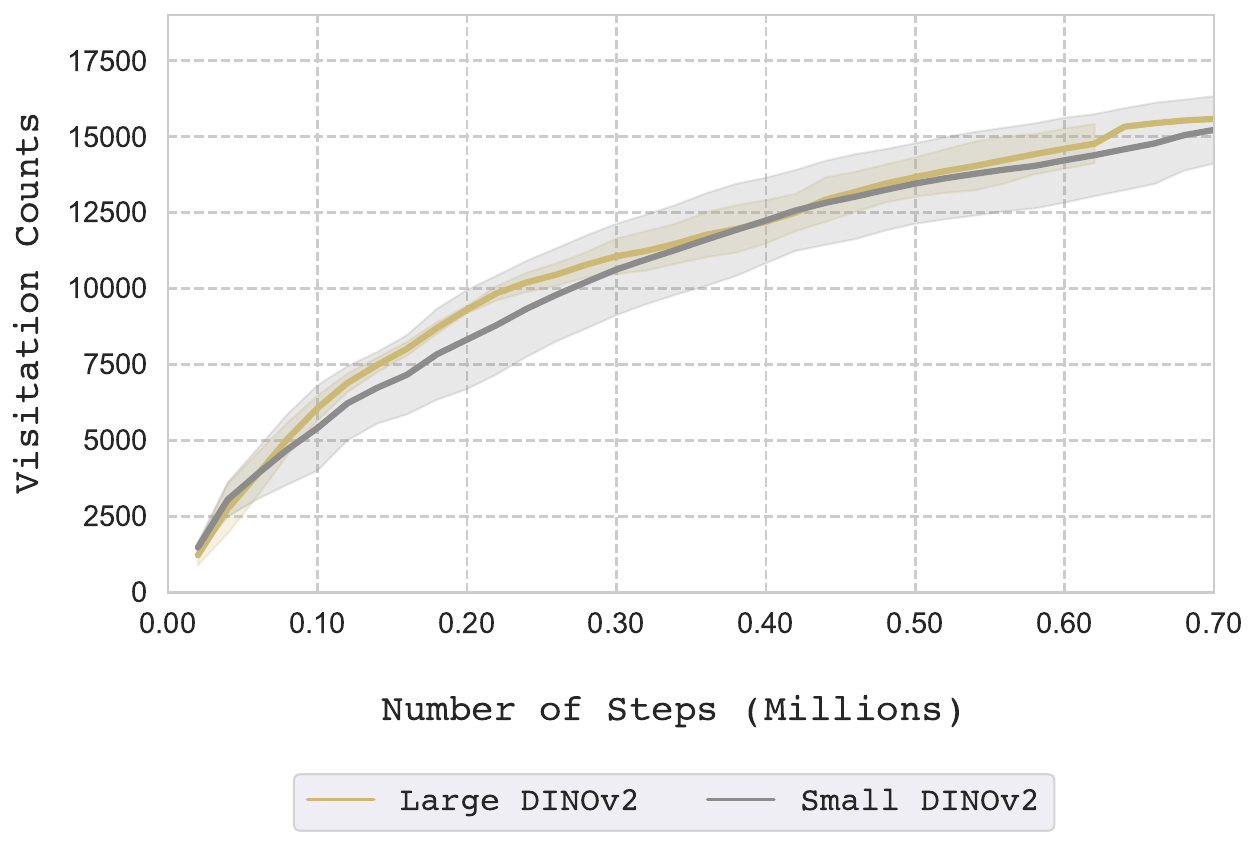}
			 };
			 \node[text width=6cm, text centered, above of=img2] at (0.3, 2.3) {\fontfamily{pcr}\selectfont\scriptsize{Ablation - DINO Model Size - Apartment-0}};	
		 \end{tikzpicture}
	 \caption{\textbf{Comparing different model sizes for DINO:} We test whether a larger model size for DINO improves performance for Habitat.
	 Our findings reveal that a larger model size does not improve performance by large margin. We argue that the model size is already large enough to capture the priors in the observations for clustering. 
	 This is also indicated by the rather small performance gap between DINOv2 small and DINOv2 large \citep{2021CaronDINO}.}
	 \label{fig:model_size_ablation}
 \end{figure*}
 \FloatBarrier
 In our experiments we use DINOv2 small \citep{2021CaronDINO} for the pre-trained representations. We test whether a larger model size for DINO improves performance for Habitat.
 We plot the visitation counts for the ``small'' and ``large'' model sizes in Figure \ref{fig:model_size_ablation}. Our findings reveal that a larger model size does not improve performance by large margin.
 The model size is likely already large enough to capture the priors in the observations for clustering. This is also indicated by the rather small performance gap between DINOv2 small and DINOv2 large \citep{2021CaronDINO}.
 However, inference time does increase substantially with larger model sizes, which is why we use DINOv2 small for our experiments.
 \FloatBarrier

 \subsection{Cluster Counts with and without episodic clustering}
 \begin{figure*}[htbp!]
	 \centering
		 \begin{tikzpicture}
			 \node[draw, thin, color=white, fill=white, rounded corners=0.0mm, inner sep=1mm] (img2){
				 \includegraphics[width=0.45\textwidth]{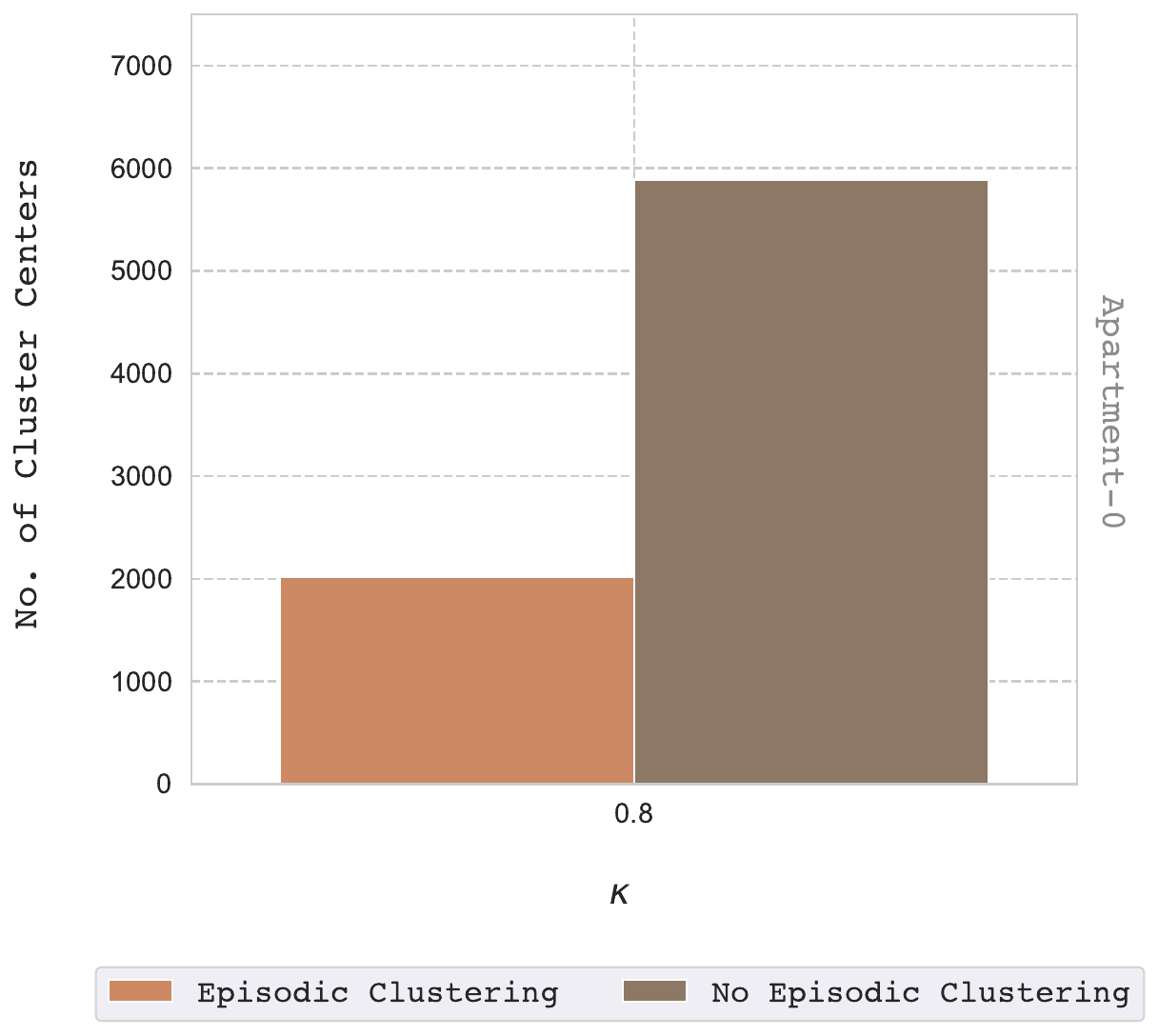}
			 };
			 \node[text width=6cm, text centered, above of=img2] at (0.3, 2.9) {\fontfamily{pcr}\selectfont\scriptsize{Ablation - Episodic Clustering Cluster Counts - Apartment-0}};	
		 \end{tikzpicture}
	 \caption{\textbf{Without episodic clustering the number of cluster centers is higher:} We plot the number of cluster centers in the cluster table with and without episodic clustering.
	 We see that without episodic clustering the number of cluster centers is higher. This is because the episodic clusters are not aggregated as well as with episodic clustering to the cluster table.}
	 \label{fig:model_size_ablation}
 \end{figure*}
 \FloatBarrier
We also investigate the effect of episodic clustering on the number of cluster centers in the cluster table. We plot the number of cluster centers in the cluster table with and without episodic clustering in Figure \ref{fig:model_size_ablation}.
We see that without episodic clustering the number of cluster centers is higher. This is because the episodic clusters are not aggregated as well as with episodic clustering to the cluster table. 
This means that not performing episodic clustering leads to a more granular cluster table, since the cluster centers in this case only represent a single embedding for every step.

\subsection{Progression of the cluster table throughout learning}
To give insight into how exploration progresses with episodic and global clustering we show the progression of the added cluster centers to the cluster table during training in Figure \ref{fig:vizdoom_env_cluster}. To locate the cluster centers within the VizDoom environment we additionally track the position of each observation and calculate the mean position of all observations in the cluster as the position for the cluster center. For visualization purposes we show the progression of the cluster table for $\kappa=0.5$

Already after 50K steps of training the first 4 are represented in the cluster table. At 500K steps almost all rooms are represented in the cluster table. This means that the intrinsic reward will be calculated from these cluster centers as reference points, effectively abstracting the state space. After convergence, the number of cluster centers barely changes since an optimal policy has been found.
\FloatBarrier
\begin{figure}[htbp!]
    \centering
	\subfigure{
		\begin{tikzpicture}
			\node[draw, inner sep=0.5mm, thin, rounded corners=1mm] at (0,0) {\includegraphics[width=.25\textwidth]{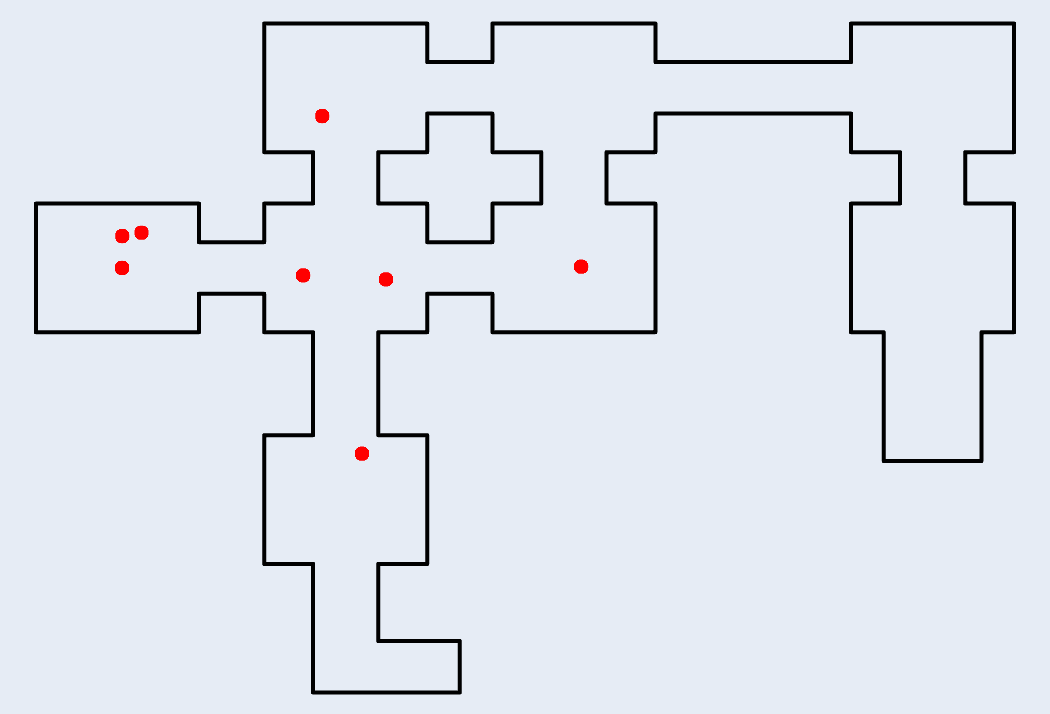}};
			\node[draw, inner sep=0.5mm, thin, rounded corners=1mm] at (4.8,0) {\includegraphics[width=.25\textwidth]{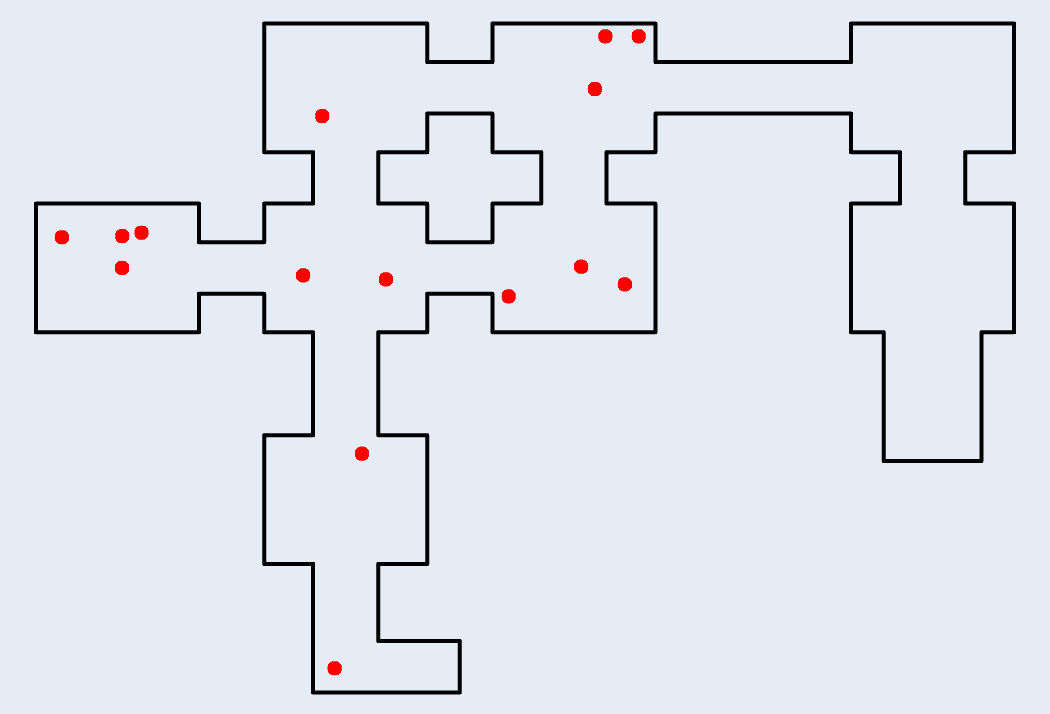}};
			\node[draw, inner sep=0.5mm, thin, rounded corners=1mm] at (9.6,0) {\includegraphics[width=.25\textwidth]{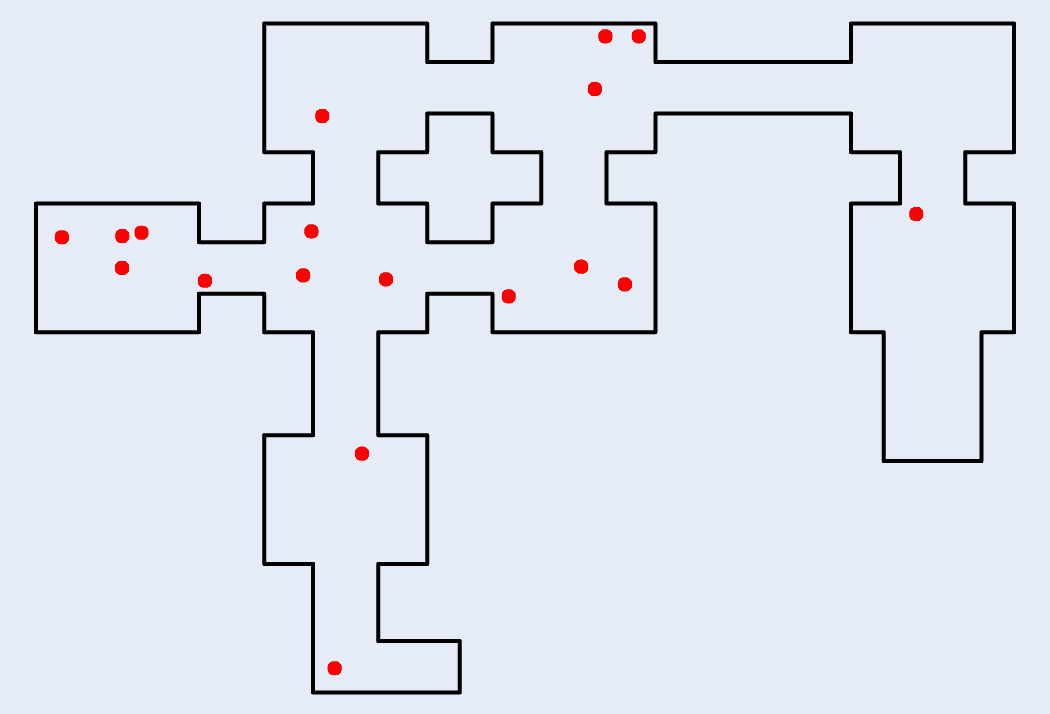}};
		
			\node[draw, inner sep=0.5mm, thin, rounded corners=1mm] at (0,-4) {\includegraphics[width=.25\textwidth]{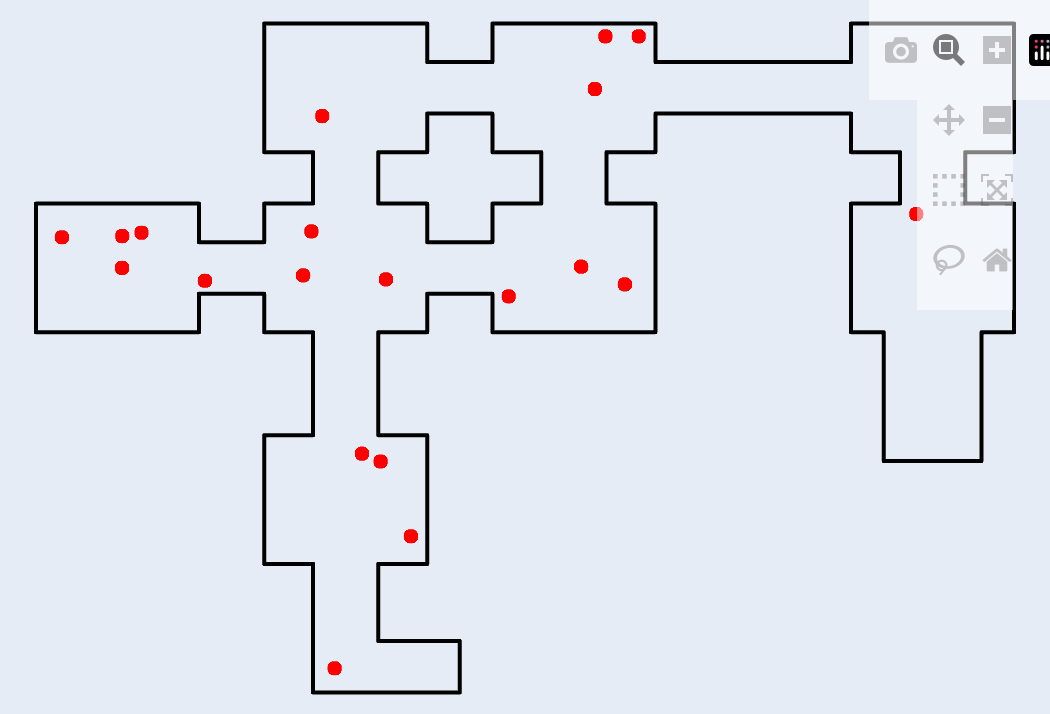}};
			\node[draw, inner sep=0.5mm, thin, rounded corners=1mm] at (4.8,-4) {\includegraphics[width=.25\textwidth]{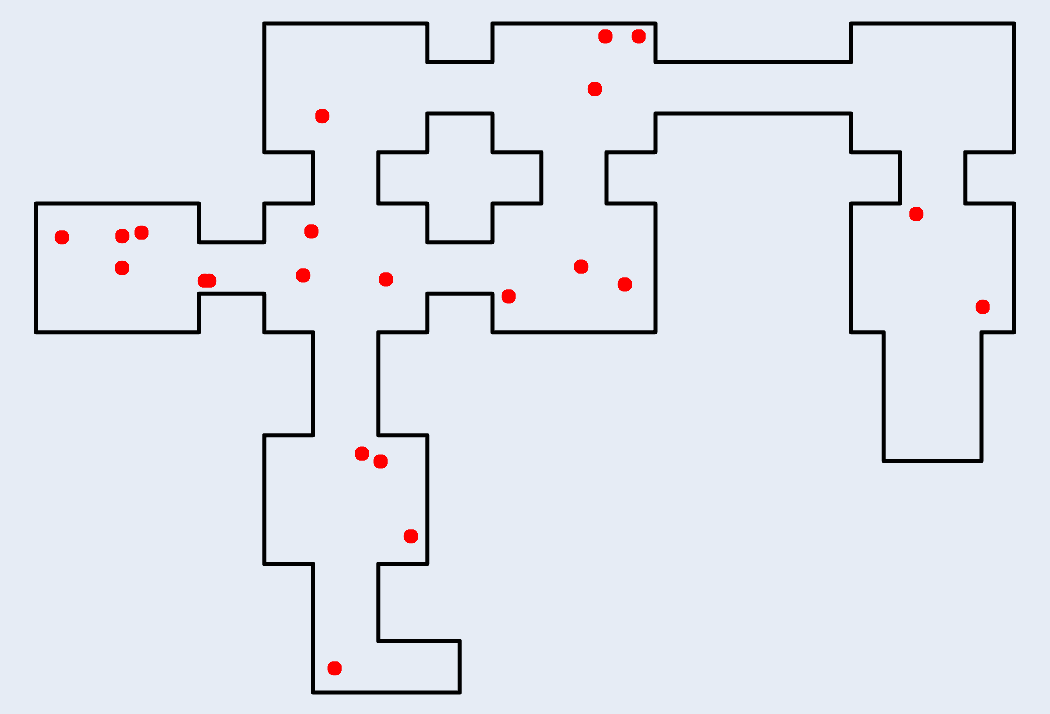}};
			\node[draw, inner sep=0.5mm, thin, rounded corners=1mm] at (9.6,-4) {\includegraphics[width=.25\textwidth]{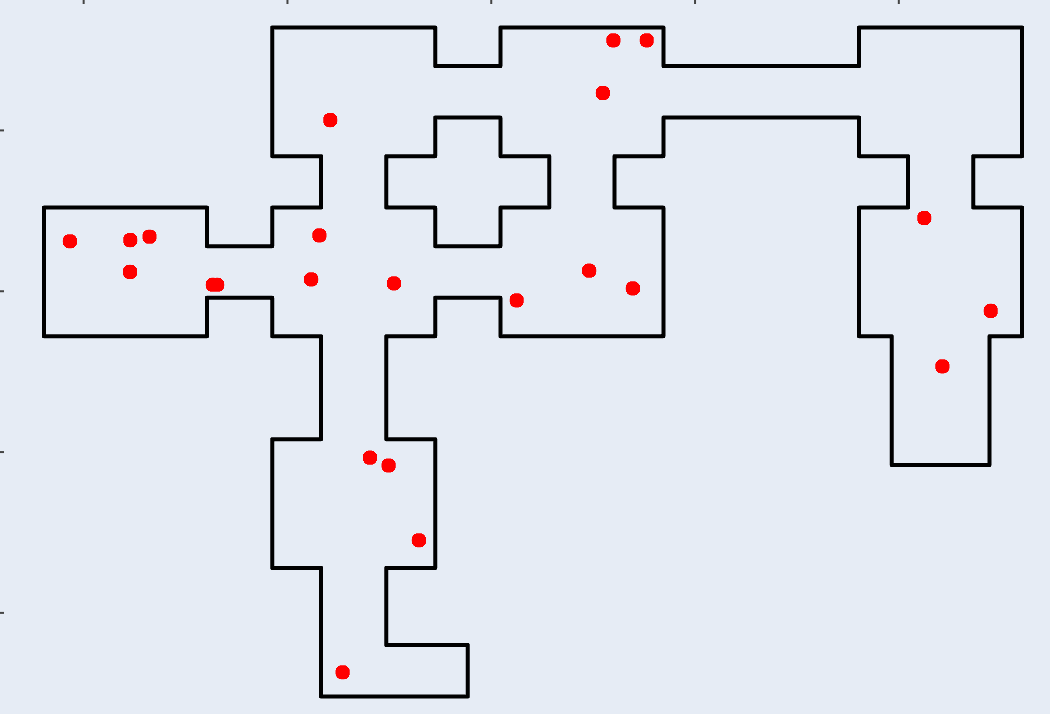}};
			
			\node[text width=4cm, text centered, above of=img1] at (0, -2.9) {50k steps};
			\node[text width=4cm, text centered, above of=img1] at (4.8, -2.9) {250k steps};
			\node[text width=4cm, text centered, above of=img1] at (9.6, -2.9) {500k steps};
			\node[text width=4cm, text centered, above of=img1] at (0, -6.9) {1M steps};
			\node[text width=4cm, text centered, above of=img1] at (4.8, -7.2) {1.6M steps \\ (before convergence)};
			\node[text width=4cm, text centered, above of=img1] at (9.6, -6.9) {5M steps};

		\end{tikzpicture}
	} 
	\caption{\textbf{Cluster centers emerge in previously unexplored areas:} We visualize the progression of the cluster table throughout learning for VizDoom.
	At the beginning of training the cluster table is empty, but as the agent explores the environment, cluster centers emerge in previously unexplored areas. Already at 50k steps the cluster table starts to fill up with cluster centers, mapping
	out the first 5 rooms of the environment. At convergence for all rooms there is at least a cluster center in the cluster table. After convergence the cluster table is stable since the optimal policy has been found.}
\label{fig:vizdoom_env}
\end{figure}
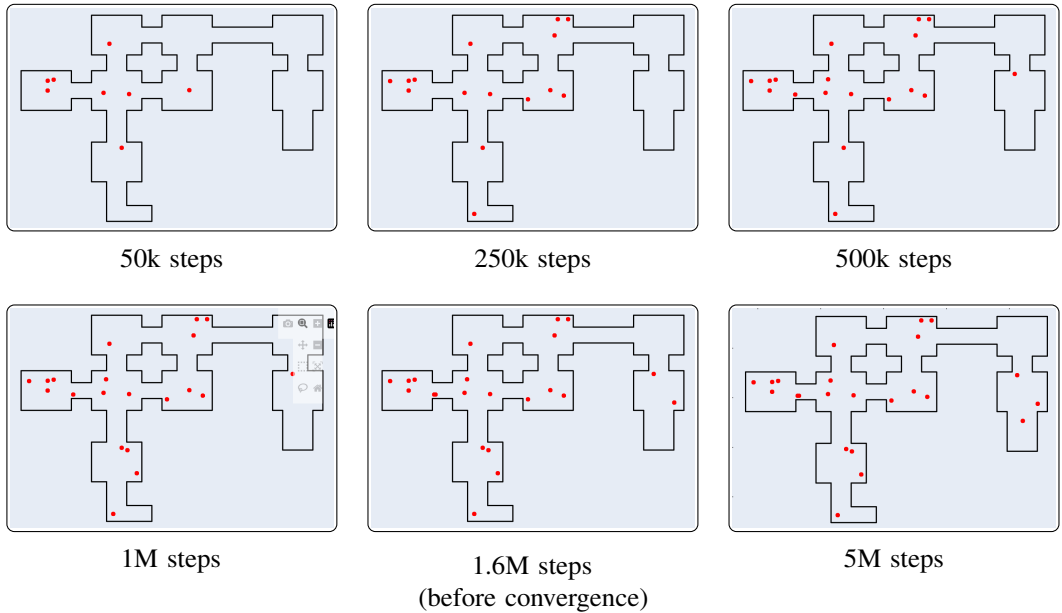
\FloatBarrier

\newpage
\section{Environments}
\label{sec:environments}
\subsection{VizDoom Environment Details}
\FloatBarrier
\begin{figure}[htbp!]
    \centering
	\subfigure{
		\begin{tikzpicture}
			\node[draw, inner sep=0.5mm, thin, rounded corners=1mm] at (0,0) {\includegraphics[width=.15\textwidth]{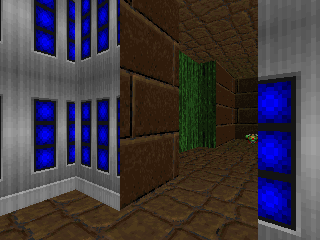}};
			\node[draw, inner sep=0.5mm, thin, rounded corners=1mm] at (2.8,0) {\includegraphics[width=.15\textwidth]{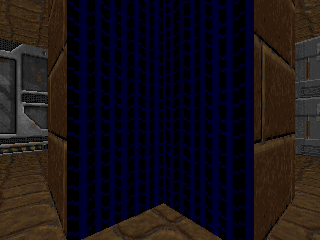}};
			\node[draw, inner sep=0.5mm, thin, rounded corners=1mm] at (5.6,0) {\includegraphics[width=.15\textwidth]{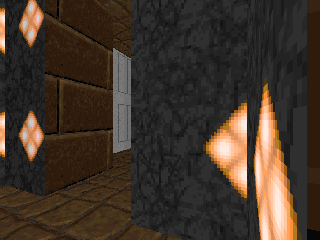}};
		
			\node[draw, inner sep=0.5mm, thin, rounded corners=1mm] at (0,-2.5) {\includegraphics[width=.15\textwidth]{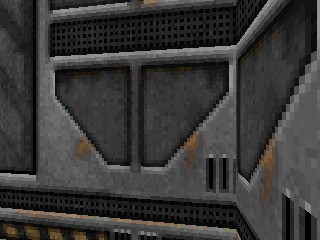}};
			\node[draw, inner sep=0.5mm, thin, rounded corners=1mm] at (2.8,-2.5) {\includegraphics[width=.15\textwidth]{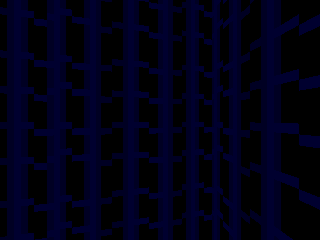}};
			\node[draw, inner sep=0.5mm, thin, rounded corners=1mm] at (5.6,-2.5) {\includegraphics[width=.15\textwidth]{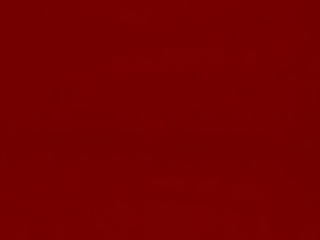}};

		\end{tikzpicture}
	} \hfil
	\subfigure{
		\includegraphics[scale=0.6]{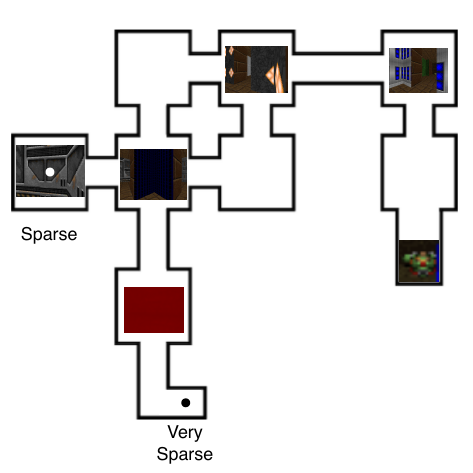}
	}
	\caption{\textbf{VizDoom:} The VizDoom environment is a 3-D environment based on the 3-D game Doom. The observations are rooms with different textures, where some textures even move.
	The agent starts in the room on the left and has to reach the armor in the room on the right. There are two configurations for this enviroments: ``Sparse'' and ``VerySparse'', which differ	
	in the starting position of the agent.}
\label{fig:vizdoom_env}
\end{figure}
\FloatBarrier

We use the ``MyWayHome'' environment from \textbf{VizDoom} \citep{Kempka2016ViZDoom} in both its sparse reward configurations ``Sparse'' and ``VerySparse''. 
In this environment the agent has to reach a piece of armor located in the farthest
accesible room of the environment. The configurations differ in the starting position of the agent, one being 
further away from the goal than the other.
In ``Sparse'', 270 steps away from goal and in ``VerySparse'', 340
steps away from goal. The agent has an action space of \emph{Wait, Left,
    Right, Forward} and an episode has a length of 2100 steps. We use a standard
frame skip of $4$. 
The agent receives a reward of $+1$ when it reaches the armor and a $-0.0001$ penalty for every
timestep.

\subsubsection{Hyperparameters-VizDoom}
For VizDoom, we show the hyperparameters for PPO in Table \ref{table:ppo_hyperparameters_vizdoom} and the hyperparameters for the episodic clustering in Table \ref{table:cosine_episodic_parameters_vizdoom}. Additionally,
we also show the hyperparameters for RECODE in Table \ref{tab:recode_parameters_vizdoom}. We use the same hyperparameters for both ``Sparse'' and ``VerySparse'' configurations. Note that in Table \ref{tab:recode_parameters_vizdoom}
we tried different sizes for the RECODE memory as we found this had the most predictable effect on performance, however for VerySparse even with different memory sizes we were not able to get RECODE to work.
\begin{table}[htbp!]
	\centering
	\begin{tabular}{|c|c|}
		\hline
		\textbf{Hyperparameter} & \textbf{Value} \\
		\hline
		Learning rate & 0.0001 \\
		\hline
		Batch size & 256 \\
		\hline
		Epochs & 4 \\
		\hline
		Gamma & 0.99 \\
		\hline
		Clip epsilon & 0.1 \\
		\hline
		Entropy coefficient & 0.005 \\
		\hline
		Value function coefficient & 0.5 \\
		\hline
		Workers & 32 \\
		\hline
		Recurrence & 64 \\
		\hline
		Intrinsic Reward Scaling & 0.1 \\
		\hline
		\end{tabular}
	\caption{PPO Hyperparameters for VizDoom}
	\label{table:ppo_hyperparameters_vizdoom}
	\end{table}

	\begin{table}[htbp!]
		\centering
		\begin{tabular}{|c|c|}
		\hline
		\textbf{Parameter} & \textbf{Value} \\
		\hline
		Cosine Similarity Threshold $\kappa$ (DINO) & 0.8 \\
		\hline
		Cosine Similarity Threshold $\kappa$ (Random) & 0.8 \\
		\hline
		Number of Clusters $M$ & 250 \\
		\hline
		\end{tabular}
		\caption{Cosine Similarity and Episodic Clustering Parameters}
		\label{table:cosine_episodic_parameters_vizdoom}
		\end{table}

		\begin{table}[htbp!]
			\centering
			\begin{tabular}{|l|l|}
			\hline
			\textbf{Parameter} & \textbf{Value} \\
			\hline
			RECODE memory size & $\{100,\ \textbf{700}, \ 1 \times 10^4 \}$ \\
			RECODE discount $\gamma$ & \textbf{0.999} \\
			RECODE insertion probability $\eta$ & \textbf{0.05} \\
			RECODE relative tolerance $\kappa$ & \textbf{0.2} \\
			RECODE reward constant $c$ & \textbf{0.01} \\
			RECODE decay rate $\tau$ & \textbf{0.9999} \\
			RECODE neighbors $k$ & \textbf{20} \\
			\hline
			\end{tabular}
			\caption{RECODE Parameters VizDoom - Bold value is used for experiments}
			\label{tab:recode_parameters_vizdoom}
			\end{table}
\FloatBarrier

\subsection{Habitat Environment Details}
\label{sec:habitat_env}
\FloatBarrier
\begin{figure}[htbp!]
    \centering
	\subfigure{
		\begin{tikzpicture}
			\node[draw, inner sep=0.5mm, thin, rounded corners=1mm] at (0,0) {\includegraphics[width=.12\textwidth]{figures/hbc/img1.png}};
			\node[draw, inner sep=0.5mm, thin, rounded corners=1mm] at (2.3,0) {\includegraphics[width=.12\textwidth]{figures/hbc/img2.png}};
			\node[draw, inner sep=0.5mm, thin, rounded corners=1mm] at (4.6,0) {\includegraphics[width=.12\textwidth]{figures/hbc/img3.png}};
		
			\node[draw, inner sep=0.5mm, thin, rounded corners=1mm] at (0,-2.3) {\includegraphics[width=.12\textwidth]{figures/hbc/emb1.png}};
			\node[draw, inner sep=0.5mm, thin, rounded corners=1mm] at (2.3,-2.3) {\includegraphics[width=.12\textwidth]{figures/hbc/emb2.png}};
			\node[draw, inner sep=0.5mm, thin, rounded corners=1mm] at (4.6,-2.3) {\includegraphics[width=.12\textwidth]{figures/hbc/emb3.png}};

		\end{tikzpicture}
	}\hfil
	\subfigure{
		\includegraphics[scale=0.3]{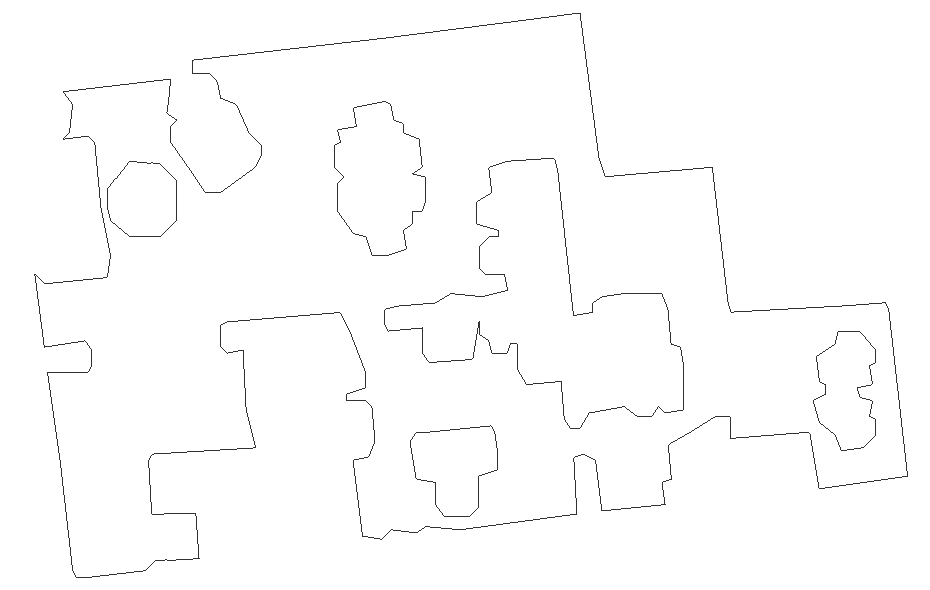}
	}
	\caption{\textbf{Habitat (Apartment-0):} The Habitat environment is a real world 3-D environment with complex observations. The observations are real world scenes of apartments, hotels and office rooms.
	The observations are very complex and thus we expect the pre-trained representations to be more effective than random features.	}
\label{fig:habitat_env}
\end{figure}
\FloatBarrier

\textbf{Habitat-Replica} \citep{habitat19iccv,szot2021habitat,puig2023habitat3} is a framework for exploration in
real world 3-D scenes. The Habitat-Lab API is made to be used with different
datasets. We use the Replica dataset \citep{replica19arxiv} as in \citet{parisi2021interesting}, where we 
train on the ``Apartment-0'' environment, which is the biggest room in the
Replica dataset. 
The agent has an action space of \emph{Forward} by 0.25
coordinate points, \emph{Left 10°} and \emph{Right 10°} and the episode has a
length of 500 steps. 
The agent is rewarded with intrinsic reward only, where the count of visited locations is the performance metric.

\subsubsection{Calculating agent position for visitation counts}
We calculate the position of the agent for the visitation counts as follows:
\[
\text{{position}} = \text{{round}}\left(\frac{{\text{{round}}(\text{{position}}, 2) \times 20}}{20}\right).
\]
This is directly taken from \cite{parisi2021interesting}'s experiments. The position is a tuppe $(x,y,z)$. We round the position to two decimal places and then multiply it by 20. 
This is done to get a more granular position, since the agent moves by 0.25 coordinate points.
Finally we round the position to the nearest integer, which is the position we use for the visitation counts. 
The visitations counts are then simply counted by aggregating them in a Python dictionary.

\subsubsection{Hyperparameters}
We show the hyperparameters for PPO in Table \ref{table:ppo_hyperparameters_habitat} and the hyperparameters for the episodic clustering in Table \ref{table:cosine_episodic_parameters_habitat}. Again, we 
also show the hyperparameters for RECODE in Table \ref{tab:recode_parameters_habitat}. For our method with random features we had to increase $\kappa = 0.9$ (compared to $\kappa = 0.8$ in VizDoom) for adequate performance.
For the DINO features $\kappa=0.8$ works well for both enviroments.
\begin{table}[htbp!]
	\centering
	\begin{tabular}{|c|c|}
	\hline
	\textbf{Hyperparameter} & \textbf{Value} \\
	\hline
	Learning rate & 0.0001 \\
	\hline
	Batch size & 256 \\
	\hline
	Epochs & 4 \\
	\hline
	Gamma & 0.99 \\
	\hline
	Clip epsilon & 0.1 \\
	\hline
	Entropy coefficient & 0.005 \\
	\hline
	Value function coefficient & 0.5 \\
	\hline
	Workers & 1 \\
	\hline
	Recurrence & 64 \\
	\hline
	Intrinsic Reward Scaling & 0.1 \\
	\hline
	\end{tabular}
	\caption{PPO Hyperparameters for Habitat}
	\label{table:ppo_hyperparameters_habitat}
	\end{table}

	\begin{table}[htbp!]
		\centering
		\begin{tabular}{|c|c|}
		\hline
		\textbf{Parameter} & \textbf{Value} \\
		\hline
		Cosine Similarity Threshold $\kappa$ (DINO) & 0.8 \\
		\hline
		Cosine Similarity Threshold $\kappa$ (Random) & 0.9 \\
		\hline
		Number of Clusters $M$ & 30 \\
		\hline
		\end{tabular}
		\caption{Cosine Similarity and Episodic Clustering Parameters}
		\label{table:cosine_episodic_parameters_habitat}
		\end{table}

		\begin{table}[htbp!]
			\centering
			\begin{tabular}{|l|l|}
			\hline
			\textbf{Parameter} & \textbf{Value} \\
			\hline
			RECODE memory size & $\{100,\ \textbf{700}, \ 1 \times 10^4, \ 5 \times 10^4\}$ \\
			RECODE discount $\gamma$ & \textbf{0.999} \\
			RECODE insertion probability $\eta$ & \textbf{0.05} \\
			RECODE relative tolerance $\kappa$ & \textbf{0.2} \\
			RECODE reward constant $c$ & \textbf{0.01} \\
			RECODE decay rate $\tau$ & \textbf{0.9999} \\
			RECODE neighbors $k$ & \textbf{20} \\
			\hline
			\end{tabular}
			\caption{RECODE Parameters Habitat - Bold value is used for the experiments.}
			\label{tab:recode_parameters_habitat}
			\end{table}
\FloatBarrier
\newpage

\section{Implementation Details}
\label{sec:implementation_details}
\subsection{Observation encoding and network architecture.}
\begin{figure*}[htbp!]
	\centering
		\begin{tikzpicture}
			\node[draw, thin, color=white, fill=white, rounded corners=0.0mm, inner sep=1mm] (img2){
				\includegraphics[width=0.5\textwidth]{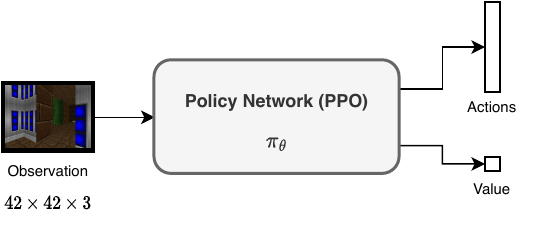}
			};
		\end{tikzpicture}
	\caption{\textbf{High level overview of policy network:} The policy network receives the observations and outputs the logits for the action probabilities and the value function.}
	\label{fig:policy_network}
\end{figure*}
As policy network we take the Proximal Policy Optimization (PPO) \citep{2017SchulmanPPO} implementation from \href{https://github.com/lcswillems/rl-starter-files}{lcswillems} and modify it to our needs.
As shown in Figure \ref{fig:policy_network}, we downscale the observations to $42 \times 42 \times 3$ and feed the downscaled observations to the following network architecture,
\begin{lstlisting}[language=Python, caption=PPO Network Architecture, label=lst:ppo_architecture]
	[...
	ACModel(
		(image_conv): Sequential(
		  (0): Conv2d(3, 32, kernel_size=(3, 3), stride=(2, 2), padding=(1, 1))
		  (1): ELU(alpha=1.0)
		  (2): Conv2d(32, 32, kernel_size=(3, 3), stride=(2, 2), padding=(1, 1))
		  (3): ELU(alpha=1.0)
		  (4): Conv2d(32, 32, kernel_size=(3, 3), stride=(2, 2), padding=(1, 1))
		  (5): ELU(alpha=1.0)
		  (6): Conv2d(32, 32, kernel_size=(3, 3), stride=(2, 2), padding=(1, 1))
		  (7): ELU(alpha=1.0)
		)
		(memory_rnn1): LSTMCell(288, 256)
		(memory_rnn2): LSTMCell(288, 256)
		(actor): Sequential(
		  (0): Linear(in_features=256, out_features=64, bias=True)
		  (1): Tanh()
		  (2): Linear(in_features=64, out_features=4, bias=True)
		)
		(critic): Sequential(
		  (0): Linear(in_features=256, out_features=64, bias=True)
		  (1): Tanh()
		  (2): Linear(in_features=64, out_features=1, bias=True)
		)
	  )

\end{lstlisting}
where the ``out\_features'' for the actor and critic are the number of actions and $1$, respectively. 
Since we have partial observability in our 3-D environments, we additionally use two LSTM cells after the convolutional encoder.

\subsection{DINO Setup}
\begin{figure*}[htbp!]
	\centering
		\begin{tikzpicture}
			\node[draw, thin, color=white, fill=white, rounded corners=0.0mm, inner sep=1mm] (img2){
				\includegraphics[width=0.5\textwidth]{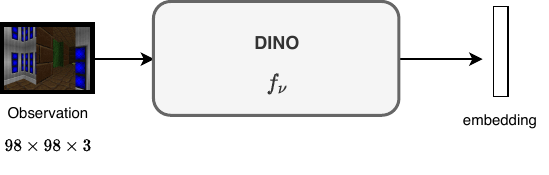}
			};
		\end{tikzpicture}
	\caption{\textbf{High level overview of how we use DINO:} We use the pre-trained DINO model to extract embeddings from the observations. We then use the cosine similarity to cluster the embeddings.}
	\label{fig:dino_network}
\end{figure*}
For the pre-trained DINO representations we use the
``small'' version of the ``dino-v2'' \citep{2023OquabDINOv2} model. 
As shown in Figure \ref{fig:dino_network}, the model receives an input of size $98 \times 98 \times 3$ since it splits the observation into $14 \times 14$ patches.
After passing the observations through the model, we take the CLS token of the DINO model as feature vector which is $384$-dimensional. 
For the model size ablation, we use the ``large'' model instead which has a $1024$-dimensional feature vector with $196 \times 196 \times 3$ sized
observations as input.

\subsection{Random Features Setup}
For the random features we use the target network in Random Network Distillation which is analogous to Listing \ref{lst:ppo_architecture} without the LSTM cells.
After the convolutional encoder we simply output a $384$-dimensional feature vector.  As a consquenece, the embeddings, i.e., the cluster means 
are also $384$-dimensional. 

\subsection{Other Libraries}
For the Gaussian mixture model we simply use the implementation from \href{https://scikit-learn.org/stable/modules/generated/sklearn.mixture.GaussianMixture.html}{scikit-learn}.

\end{document}